\documentclass[10pt,twocolumn]{article} 

\usepackage{OptimLabTwoColumn}

\usepackage{mathtools}
\usepackage{algorithm}
\usepackage{algpseudocode}
\usepackage{amsmath,amssymb,amsthm,amsfonts}
\usepackage{booktabs}
\usepackage{subfigure}
\usepackage{url}
\usepackage{makecell}
\usepackage{multirow}
\usepackage[symbol]{footmisc}
\usepackage{tikz}
\usepackage{caption}
\usepackage{hyperref}
\usepackage{bbm}

\usetikzlibrary{shapes,decorations,arrows,calc,arrows.meta,fit,positioning}
\tikzset{
    -Latex,auto,node distance =1 cm and 1 cm,semithick,
    state/.style ={ellipse, draw, minimum width = 0.7 cm},
    point/.style = {circle, draw, inner sep=0.04cm,fill,node contents={}},
    bidirected/.style={Latex-Latex,dashed},
    el/.style = {inner sep=2pt, align=left, sloped}
}

\def\cL{{\cal L}}

\def\cX{{\cal X}}

\def\cL{{\cal L}}

\newcommand{\bz}{{\bf z}}

\newcommand{\bX}{{\bf X}}
\newcommand{\bx}{{\bf x}}

\newcommand{\mbR}{\mathbb{R}}
\newcommand{\mbE}{\mathbb{E}}

\newcommand{\bc}{\begin{center}}
\newcommand{\ec}{\end{center}}
\newcommand{\be}{\begin{equation}}
\newcommand{\ee}{\end{equation}}
\newcommand{\ba}{\begin{array}}
\newcommand{\ea}{\end{array}}
\newcommand{\bean}{\begin{eqnarray*}}
\newcommand{\eean}{\end{eqnarray*}}
\newcommand{\bea}{\begin{eqnarray}}
\newcommand{\eea}{\end{eqnarray}}
\newcommand{\ben}{\begin{enumerate}}
\newcommand{\een}{\end{enumerate}}
\newcommand{\bed}{\begin{itemize}}
\newcommand{\eed}{\end{itemize}}

\newtheorem{assumption}{Assumption}

\begin{document}

\title{Joint Distributional Learning via Cramer-Wold Distance}

\author{
  Seunghwan An \textnormal{and} Jong-June Jeon\thanks{Corresponding author.} \\
  Department of Statistical Data Science, University of Seoul, S. Korea \\
  \texttt{\{dkstmdghks79, jj.jeon\}@uos.ac.kr} \\
}

\maketitle
\thispagestyle{empty}

\begin{abstract}
The assumption of conditional independence among observed variables, primarily used in the Variational Autoencoder (VAE) decoder modeling, has limitations when dealing with high-dimensional datasets or complex correlation structures among observed variables. To address this issue, we introduced the Cramer-Wold distance regularization, which can be computed in a closed-form, to facilitate joint distributional learning for high-dimensional datasets. Additionally, we introduced a two-step learning method to enable flexible prior modeling and improve the alignment between the aggregated posterior and the prior distribution. Furthermore, we provide theoretical distinctions from existing methods within this category. To evaluate the synthetic data generation performance of our proposed approach, we conducted experiments on high-dimensional datasets with multiple categorical variables. Given that many readily available datasets and data science applications involve such datasets, our experiments demonstrate the effectiveness of our proposed methodology.
\end{abstract}

\section{Introduction}
\label{sec:1}


The Variational Autoencoder (VAE) is a generative model utilized to estimate the underlying distribution of a given dataset \cite{Kingma2014, JimenezRezende2014StochasticBA}. The primary objective of VAE is to maximize the Evidence Lower Bound (ELBO) of the observation $\bx$, thereby enabling the generative model to produce synthetic data that closely resembles the observed dataset. Note that the generative model of VAE is written as follows:
\bea \label{eq:gen_model}
\int p(\bz) p(\bx|\bz) d\bz,
\eea
where $p(\mathbf{z})$ represents the prior distribution of the latent variable $\bz$ and $p(\mathbf{x}|\mathbf{z})$ corresponds to the decoder. 

The ELBO is derived as a lower bound on the log-likelihood of an individual observation $\bx$, making it a local approximation for that specific data point. To achieve equality in the ELBO for accurately recovering the given observation, the Kullback-Leibler (KL) divergence between the proposal posterior $q(\bz|\bx)$ and the true posterior $p(\bz|\bx)$ distributions should be minimized, ideally reaching zero. This means that the proposal posterior distribution should have an infinite capacity, ensuring that the generative model can generate the synthetic data accurately.

However, conventional VAE approaches typically assume that a prior distribution follows a standard Gaussian distribution. This choice offers certain advantages, such as having a closed-form KL-divergence and improved sampling efficiency. However, it also implies that this prior has a `finite' capacity. Consequently, the aggregated posterior \cite{pmlr-v84-tomczak18a}, denoted as $\int q(\mathbf{z}|\mathbf{x}) p(\mathbf{x}) d\mathbf{x}$, can significantly differ from the prior. This deviation in the distributions carries a notable implication: the generative model, as represented by \eqref{eq:gen_model}, can \textit{not} effectively produce synthetic data that closely resembles the original data because the decoder $p(\bx|\bz)$ is trained using latent variables sampled from the proposal posterior $q(\bz|\bx)$.


Hence, generating high-quality synthetic datasets crucially depends on aligning the aggregated posterior with the chosen prior distribution \cite{Bousquet2017FromOT}. This alignment process involves parameterizing the prior using trainable parameters. Several previous studies, including \cite{pmlr-v84-tomczak18a, guo2020variational, hajimiri2021semi}, have parameterized the prior using a mixture (Gaussian) distribution. However, when maximizing the ELBO jointly with this complex prior, it often necessitates intricate mathematical derivations like the density ratio trick \cite{takahashi2019variational} and the application of a greedy algorithm \cite{pmlr-v84-tomczak18a}.

Within this context, several studies such as \cite{Li2015GenerativeMM, Sbakan2020OnTE, pmlr-v80-achlioptas18a, dai2018diagnosing, NIPS2017_7a98af17, engel2018latent}, have adopted the \textit{two-step learning} method. This approach involves a separate training process where the alignment of the prior with the aggregated posterior is carried out independently. It not only helps alleviate the challenges associated with intricate derivations but also provides a more stable training process. Additionally, it enables the construction of a flexible learning pipeline \cite{Sbakan2020OnTE}.

In this paper, we introduce a novel framework for two-step learning that leverages the distributional learning of VAE \cite{An2023DistributionalLO}. Our specific focus is on applying this framework to datasets characterized by high dimensionality and containing multiple categorical attributes. We emphasize this for two compelling reasons.

(\textit{high-dimensional}) Firstly, when dealing with high-dimensional data, it is a common practice to assume conditional independence among observed variables given the latent variable. However, this assumption can pose challenges when the dimensionality of the latent variable is smaller than that of the observations, causing inaccuracies in capturing this conditional independence. We employ two regularization techniques to tackle this issue and effectively model the correlation structure among observed variables. These regularizations are based on the Cramer-Wold distance \cite{Tabor2018CramerWoldA, KNOP2022119} and classification loss \cite{pmlr-v157-zhao21a, Park2018DataSB}. Notably, the Cramer-Wold distance shares similarities with the sliced-Wasserstein distance but offers the advantage of having a closed-form solution.

(\textit{multiple-categorical}) In many readily available public datasets and data science applications, it's common to encounter datasets that include categorical variables \cite{Camino2018GeneratingMS}. For example, among the entire dataset collection available at \url{archive.ics.uci.edu}, datasets that contain categorical and mixed columns account for approximately 65.4\% of the total. Hence, we conduct experiments on publicly available real tabular datasets that consist of multiple categorical variables. These experiments showcase the excellent performance of our proposed model in generating synthetic data.


Our paper makes two primary contributions, which can be summarized as follows:
\ben
    \item We present a novel framework for two-step learning and provide both theoretical and numerical distinctions from existing methods within this category.
    \item We specifically utilize the Cramer-Wold distance to enable joint distributional learning for multiple-categorical datasets and demonstrate its effectiveness through a series of numerical experiments.
\een

\section{Related Work}
\label{sec:2}

\textbf{Distributional learning (generative modeling).}
Distributional learning involves the estimation of the underlying distribution of an observed dataset. Generative models based on latent spaces aim to perform distributional learning by generating data closely resembling a given dataset. An early and prominent example of generative modeling is the VAE. However, VAE faced limitations in generative performance due to a misalignment between the distribution of representations that learned information from the observations, known as the aggregated posterior, and the prior distribution.

To address this issue, \cite{Makhzani2015AdversarialA} introduced the Adversarial AutoEncoder (AAE), which directly minimizes the aggregated posterior and prior using the adversarial loss from the GAN framework. This introduction of adversarial loss made it easier to compute, even when the aggregated posterior took complex forms, unlike KL-divergence. Similarly, \cite{Bousquet2017FromOT} proposed the penalized optimal transport (POT). The POT's objective function consists of a reconstruction loss (cost) and a penalty term that minimizes the divergence (distance) between the aggregated posterior and prior distributions. Subsequent research incorporated various divergences into this penalty term, such as Maximum Mean Discrepancy (MMD) \cite{Tolstikhin2017WassersteinA}, Sliced-Wasserstein distance \cite{Deshpande2018GenerativeMU}, and Cramer-Wold distance \cite{Tabor2018CramerWoldA}. Notably, Sliced-Wasserstein and Cramer-Wold distances are based on random projections of high-dimensional datasets onto one-dimensional subspaces, resolving challenges in calculating distances between multivariate distributions.

While these methods commonly utilize divergence for alignment in the latent space, some studies directly introduce divergence into the data space. For instance, \cite{10.5555/3020847.3020875, Li2015GenerativeMM, Hofert2021RafterNetPP} employed MMD, and \cite{Deshpande2018GenerativeMU} used the sliced Wasserstein distance for reconstruction error. More recently, \cite{An2023DistributionalLO} introduced the continuous ranked probability score (CRPS), a proper scoring rule that measures the distance between the proposed cumulative distribution function (CDF) and the ground-truth CDF of the underlying distribution. It shows theoretically that it is feasible to minimize the KL-divergence between the ground-truth density and the density estimated through generative modeling.

\textbf{Two-step learning.}
As previously discussed, within the VAE framework, optimizing the ELBO while simultaneously learning both the decoder and complex prior parameters often involves complex mathematical derivations, such as the density ratio trick \cite{takahashi2019variational}, and a greedy algorithm \cite{pmlr-v84-tomczak18a}. The requirement for a closed-form expression of the ELBO has limited the exploration of new approaches to modeling priors.

However, \cite{Sbakan2020OnTE} has revealed that two-step training can be thought of as a simple combination of existing methods for fitting the decoder and prior model. This approach offers the added benefit of flexibility in the learning process, allowing for straightforward adjustments to the prior modeling when the necessary method for learning these distributions is available.

\cite{Li2015GenerativeMM, Sbakan2020OnTE, pmlr-v80-achlioptas18a} employed an AutoEncoder to fit the decoder, while \cite{dai2018diagnosing, NIPS2017_7a98af17, engel2018latent} used the VAE framework in the first step of training. A common theme in these papers was the learning of the prior distribution in the second step to align with the aggregated posterior (distribution of representations) \cite{Makhzani2015AdversarialA}. Notably, \cite{dai2018diagnosing} theoretically demonstrated that under the assumption that observations exist on a simple Riemannian manifold, two-step learning can approximate the ground-truth measure.

\textbf{Handing multiple-categorical datasets.}
To train the generator and discriminator networks with multiple-categorical (discrete) variables, \cite{Choi2017GeneratingMD} proposes a combination of an AutoEncoder and a GAN, which is based on \cite{Zhao2017AdversariallyRA}. The AutoEncoder directly learns from high-dimensional discrete variables, while the GAN generates the continuous latent variables of the AutoEncoder's latent space. In other words, the GAN learns the distribution of the representation vectors. Subsequent studies have adopted this approach \cite{Torfi2019GeneratingSH, Yang2019GroupedCG, Lee2020GeneratingSE, cite-key}. \cite{cite-key} employs the VAE instead of the autoencoder. On the other hand, \cite{Camino2018GeneratingMS} proposes another approach that avoids backpropagating through discrete samples by adopting the Gumbel-Softmax \cite{jang2017categorical} to make sampling from discrete distributions differentiable. Further, \cite{Kuo2022TheHG, Yale2020GenerationAE} incorporate the Wasserstein GAN with gradient penalty (WGAN-GP, \cite{pmlr-v70-arjovsky17a}) to enhance training stability and accommodate various variables types.

\textbf{Synthetic data generation.}
The synthetic data generation task actively adopts the GAN framework, as it allows for nonparametric synthetic data generation \cite{Choi2017GeneratingMD, Park2018DataSB, NEURIPS2019_254ed7d2, pmlr-v157-zhao21a, Park2021SynthesizingIC, Kuo2022TheHG, Kuo2022GeneratingSC, Yale2020GenerationAE, Hernandez2022SyntheticDG}. In particular, \cite{NEURIPS2019_254ed7d2, pmlr-v157-zhao21a} assume that continuous columns in tabular datasets can be approximated using Gaussian mixture distributions and model their decoder accordingly. They also employ the Variational Gaussian mixture model \cite{Blei2016VariationalIA}, known as \textit{mode-specific normalization}, to preprocess the continuous variables. However, this preprocessing step requires additional computational resources and hyperparameter tuning to determine the number of modes. Alternatively, other approaches proposed by \cite{Park2018DataSB, pmlr-v157-zhao21a} focus on regularizing the difference between the first and second-order statistics of the observed and synthetic datasets. 

\textbf{Correlation structure learning.}
Several studies have focused on capturing the correlation structure between variables to improve the quality of synthetic data. For instance, \cite{Yang2019GroupedCG} maximizes the correlation between two different latent vectors representing diseases and drugs. Similarly, \cite{Kuo2022TheHG} introduces an alignment loss based on the $L_2$ distance between correlation matrices. On the other hand, \cite{Torfi2019GeneratingSH} modifies the Multilayer Perceptron (MLP) with Convolutional Neural Networks (CNN).

\textbf{The learning of prior.} 
\cite{hoffman2016elbo, Makhzani2015AdversarialA} have demonstrated that the aggregated posterior is the optimal prior, which maximizes the objective function of the VAE, but it can lead to overfitting. To address this issue, \cite{tomczak2018vae, guo2020variational} proposed approximating the optimal prior by using a finite mixture of posterior distributions with trainable pseudo-inputs. However, the performance of VampPrior \cite{tomczak2018vae} is sensitive hyperparameters, such as the number of mixture components \cite{takahashi2019variational}. \cite{Makhzani2015AdversarialA, takahashi2019variational} employed the adversarial training method to regularize the VAE model by aligning the aggregated posterior with the prior distribution. 

\section{Proposal}
\label{sec:3}

Let $\bx \in \cX \subset \mbR^p$ be an observation consisting of discrete variables. $T_j$ denotes the number of levels for the discrete variables $\bx_j$ where $j \in \{1,\cdots,p\}$. We denote the ground-truth underlying distribution (probability density function, PDF) as $p(\bx)$. The decoder, posterior, and prior distributions are denoted as $p(\bx|\bz;\theta)$, $q(\bz|\bx;\phi)$, and $p(\bz;\eta)$, respectively, where $\theta, \phi, \eta$ are trainable neural network parameters. Note that the prior distribution is not fixed and is parameterized with a trainable parameter $\eta$. The aggregated posterior \cite{Makhzani2015AdversarialA, tomczak2018vae} is defined as 
\bean
q(\bz;\phi) \coloneqq \int p(\bx) q(\bz|\bx;\phi) d\bx.
\eean
Equipped with the proposal distributions above, the generative model is defined as
\bea \label{eq:latent_model}
\hat{p}(\bx;\theta,\eta) \coloneqq \int p(\bx|\bz;\theta) \cdot p(\bz;\eta) d\bz,
\eea
and it is also referred to as the estimated density function. 

Then, our primary objective is to approximate the ground-truth density by minimizing some divergence between the estimated and the ground-truth density functions. We employ the forward KL-divergence, $\mathcal{KL}(p(\bx) \| \hat{p}(\bx;\theta,\eta))$, as it is one of the most popular choices \cite{Papamakarios2019NormalizingFF}. As shown in \cite{Sbakan2020OnTE}, $\mathcal{KL}(p(\bx) \| \hat{p}(\bx;\theta,\eta))$ for the two-step learning \cite{Sbakan2020OnTE} can be re-written as 
\bea \label{eq:primary}
&& \mathcal{KL}\Big(p(\bx) \| \hat{p}(\bx;\theta,\eta)\Big) \nonumber\\
= && \underbrace{\mathcal{KL}\Big(p(\bx) q(\bz|\bx;\phi) \| p(\bx|\bz;\theta) q(\bz;\phi)\Big)}_{(i)} \nonumber\\
&+& \underbrace{\mathcal{KL}(q(\bz;\phi) \| p(\bz;\eta))}_{(ii)}
\eea
(see Appendix \ref{app:twostep} for the detailed derivation of \eqref{eq:primary}).

In \eqref{eq:primary}, the terms $(i)$ and $(ii)$ represent the objectives of training steps 1 and 2, respectively. By the distributional learning of VAE \cite{An2023DistributionalLO}, we can minimize $\mathcal{KL}(q(\bz;\phi) \| p(\bz;\eta))$, because the parameter $\phi$ is fixed during the training process of step 2. Therefore, our proposal method is mainly focused on the training process of step 1, and the differences between the existing two-step learning methods will be addressed in the following sections.

\subsection{Step 1}

The objective of step 1 training process is the term $(i)$ of \eqref{eq:primary}, and it can be re-written as
\bea \label{eq:stage1}
\cL(\theta, \phi) \coloneqq && \mathcal{KL}\Big(p(\bx) q(\bz|\bx;\phi) \| p(\bx|\bz;\theta) q(\bz;\phi)\Big) \nonumber\\
= && \mbE_{p(\bx) q(\bz|\bx;\phi)} [- \log p(\bx|\bz;\theta)] \nonumber\\
&+& \mbE_{p(\bx)} [\mathcal{KL}(q(\bz|\bx;\phi) \| q(\bz;\phi))] \nonumber\\
&-& H(p(\bx)),
\eea
where $H(\cdot)$ is the entropy function. We minimize \eqref{eq:stage1} with respect to $\theta,\phi$. The third term of \eqref{eq:stage1} of RHS is the entropy of the ground-truth density function and is constant.

\begin{assumption} \label{assump:ae}
$\bx_1,\cdots,\bx_p$ are mutually independent given $\bz$.
\end{assumption}

Our proposed model assumes that $p(\bx)$ is parametrized by a mixture of categorical distributions, i.e., the decoder $p(\bx|\bz;\theta)$ of the generative model \eqref{eq:latent_model} is defined as follows:
\bea \label{eq:ae_decoder}
p(\bx|\bz;\theta) &=& \prod_{j=1}^p p(\bx_j|\bz;\theta_j) \nonumber\\
&=& \prod_{j=1}^p \prod_{l=1}^{T_j} \pi_l(\bz;\theta_j)^{\mathbb{I}(\bx_j = l)},
\eea
by Assumption \ref{assump:ae}, where $\theta = (\theta_1,\cdots,\theta_p)$, $\pi(\cdot; \theta_j): \mbR^d \mapsto \Delta^{T_j-1}$ is a neural network parameterized with $\theta_j$, where $\Delta^{T_j-1}$ is the standard $(T_j-1)$-simplex for all $\bz \in \mbR^d$, and the subscript $l$ referes to the $l$th element of the output $\pi$. 

Then, the reconstruction loss of step 1, the first term of \eqref{eq:stage1}, is written as follows:
\bean
\mbE_{p(\bx) q(\bz|\bx;\phi)} \left[ - \sum_{j=1}^p \sum_{l=1}^{T_j} \mathbb{I}(\bx_j = l) \cdot \log \pi_l(\bz;\theta_j) \right],
\eean
which is equivalent to the cross-entropy (classification loss) for a dataset with categorical variables. 

For the computation of $\mbE_{p(\bx)} [\mathcal{KL}(q(\bz|\bx;\phi) \| q(\bz;\phi))]$, which is the second term of \eqref{eq:stage1}, the log-likelihood of the posterior distribution need to be tractable. Therefore, we parameterized the posterior with the multivariate Gaussian distribution. The posterior distribution is defined as $q(\bz|\bx;\phi) \coloneqq \mathcal{N}\big(\bz | \mu(\bx;\phi), diag(\sigma^2(\bx;\phi))\big)$, where $\mu:\mbR^{p} \mapsto \mbR^d$, $\sigma^2:\mbR^{p} \mapsto \mbR_+^d$ are neural networks parameterized with $\phi$, and $diag(a), a \in \mbR^d$ denotes a diagonal matrix with diagonal elements $a$.

However, since the second term of \eqref{eq:stage1} is still not tractable due to the presence of the aggregated posterior, we minimize the upper bound of the second term of \eqref{eq:stage1}, which is derived as follows:
\bea \label{eq:upper_entropy}
0 &\leq& \mbE_{p(\bx)} [\mathcal{KL}(q(\bz|\bx;\phi) \| q(\bz;\phi))] \\
&=& I(\bx,\bz;\phi) \nonumber\\
&\leq& \mbE_{p(\bx) q(\bz|\bx;\phi)} [\log q(\bz|\bx;\phi)] - \mbE_{p(\bx) q(\bz;\phi)} [\log q(\bz|\bx;\phi)] \nonumber\\
&=& \mbE_{p(\bx)} [-H(q(\bz|\bx;\phi))] - \mbE_{p(\bx) q(\bz;\phi)} [\log q(\bz|\bx;\phi)] \nonumber
\eea
(the detailed derivation is shown in Appendix \ref{app:twostep}).
\eqref{eq:upper_entropy} regularize the entropy of the posterior distribution and minimizing \eqref{eq:upper_entropy} means that all latent variables should not have specific information about a certain observation of $\bx$. In this paper, we will refer to this upper bound as the `entropy regularization term.'

\subsubsection{Joint Distributional Learninig}

However, when we model the decoder based on Assumption \ref{assump:ae}, it offers computational efficiency. Still, \eqref{eq:ae_decoder} struggles to capture the joint relationships among the observed variables effectively. This limitation becomes evident when dealing with a restricted latent space, where latent variables may fail to capture the conditional independence among observed variables. Consequently, the model excels at capturing only the marginal distribution of the observed dataset, lacking in capturing intricate dependencies.

To address this limitation, we employ two regularization strategies: Cramer-Wold distance regularization \cite{Tabor2018CramerWoldA, KNOP2022119} and classification loss regularization \cite{pmlr-v157-zhao21a, Park2018DataSB}. Cramer-Wold distance, while similar in spirit to the sliced-Wasserstein distribution, stands out due to its closed-form solution.


For the classification loss regularization, we first define the conditional distributions, $p(\bx_j|\bx_{-j};\varphi_j)$, which are assumed to be categorical distributions where $\bx_{-j}$ denotes the vector of $\bx$ except for $\bx_j$ for $j \in \{1,\cdots,p\}$, and $\varphi = (\varphi_1, \cdots, \varphi_p)$. And we pre-train the one-vs-all classifiers $p(\bx_j|\bx_{-j};\varphi_j)$ as follows:
\bea \label{eq:pretrain_classifier}
\max_\varphi \mbE_{p(\bx)} \left[ \sum_{j=1}^p \sum_{l=1}^{T_j} \mathbb{I}(\bx_j = l) \cdot \log p(\bx_j|\bx_{-j};\varphi_j) \right].
\eea

Finally, our objective function for the step 1 is minimizing 
\bea \label{eq:final_obj}
&-& \mbE_{p(\bx) q(\bz|\bx;\phi)} \left[ \sum_{j=1}^p \sum_{l=1}^{T_j} \mathbb{I}(\bx_j = l) \cdot \log \pi_l(\bz;\theta_j) \right] \\
&+& \mbE_{p(\bx) q(\bz|\bx;\phi)} [\log q(\bz|\bx;\phi)] - \mbE_{p(\bx) q(\bz;\phi)} [\log q(\bz|\bx;\phi)] \nonumber\\
&+& \lambda \cdot \int_{S_p} \| \mbox{sm}_{\kappa}(v^\top \bX) - \mbox{sm}_{\kappa}(v^\top \hat{\bX}) \|_2^2 d \sigma_p(v) \nonumber\\
&-& \gamma \cdot \mbE_{q(\hat{\bx};\phi,\theta)} \left[ \sum_{j=1}^p \sum_{l=1}^{T_j} \mathbb{I}(\hat{\bx}_j = l) \cdot \log p(\hat{\bx}_j|\hat{\bx}_{-j};\varphi^*_j) \right] \nonumber
\eea
with respect to $\theta, \phi$, where $\lambda \geq 0$, $\gamma \geq 0$, $S_p$ denotes the unit sphere in $\mbR^p$, $\sigma_p$ is the normalized surface measure on $S_p$, and $\varphi^*$ is the pre-trained parameter which is fixed during the training process. Also, we denote $\bX \coloneqq \{\bx_i\}_{i=1}^n$ and $\hat{\bX} \coloneqq \{\hat{\bx}_i\}_{i=1}^n$, where $\bx_i \sim p(\bx)$, $\hat{\bx}_i \sim q(\hat{\bx};\phi,\theta)$, and
\bean
q(\hat{\bx};\phi,\theta) := \int p(\bx) q(\bz|\bx;\phi) p(\hat{\bx}|\bz;\theta) d\bz d\bx.
\eean 
With a Gaussian kernel $N(\cdot, \kappa)$, the smoothen distribution is defined as $\mbox{sm}_{\kappa}(R) := \frac{1}{n} \sum_{i=1}^n N(r_i, \kappa)$, where the sample $R = \{ r_i \}_{i=1}^n$ and $r_i \in \mbR$. Furthermore, in Section \ref{sec:tabular}, we experimentally demonstrated the impact of the entropy regularization term on the synthetic data generation performance, as well as the influence of each regularization.

\subsubsection{Comparison to Prior Works}

In this section, we will demonstrate that the step 1 objective function of existing two-step learning methods differs from the term $(i)$ of \eqref{eq:primary}, which is the objective function of our step 1 training process. In short, the second term of \eqref{eq:stage1}, $\mbE_{p(\bx)} [\mathcal{KL}(q(\bz|\bx;\phi) \| q(\bz;\phi))]$, is \textit{not} minimized with existing two-step learning methods. Instead, a vanilla AutoEncoder \cite{Li2015GenerativeMM, Sbakan2020OnTE, pmlr-v80-achlioptas18a} or a VAE \cite{dai2018diagnosing, NIPS2017_7a98af17, engel2018latent} is trained. Furthermore, we also observe that minimizing the second term of \eqref{eq:stage1} is important for the model's synthetic data generation performance, as shown in Section \ref{sec:tabular}.

\textbf{AutoEncoder.} If a vanilla AutoEncoder is trained in step 1, then the objective function $\cL^{(1)}(\theta, \phi)$ can be written as:
\bean
\min_{\theta,\phi} \cL^{(1)}(\theta, \phi) \coloneqq \mbE_{p(\bx) q(\bz|\bx;\phi)} [- \log p(\bx|\bz;\theta)].
\eean
Since $-H(p(\bx))$ is constant and the second term of \eqref{eq:stage1} is always non-negative, $\cL^{(1)}(\theta, \phi) \leq \cL(\theta, \phi)$ and it means that the two-step learning method with AutoEncoder \cite{Li2015GenerativeMM, Sbakan2020OnTE, pmlr-v80-achlioptas18a} minimizes the lower bound of \eqref{eq:stage1}.

\textbf{VAE.} For the two-step learning with VAE, the objective function $\cL^{(2)}(\theta, \phi)$ is:
\bean
\min_{\theta,\phi} \cL^{(2)}(\theta, \phi) \coloneqq && \mbE_{p(\bx) q(\bz|\bx;\phi)} [- \log p(\bx|\bz;\theta)] \\
&+& \mbE_{p(\bx)} [\mathcal{KL}(q(\bz|\bx;\phi) \| p(\bz;\eta))].
\eean
And the second term of \eqref{eq:stage1} can be written as:
\bean
&& \mbE_{p(\bx)} [\mathcal{KL}(q(\bz|\bx;\phi) \| q(\bz;\phi))] \\
&=& \iint p(\bx) q(\bz|\bx;\phi) \log \frac{q(\bz|\bx;\phi)}{q(\bz;\phi)} \cdot \frac{p(\bz;\eta)}{p(\bz;\eta)} d\bx d\bz \\
&=& \mbE_{p(\bx)} [\mathcal{KL}(q(\bz|\bx;\phi) \| p(\bz;\eta))] + \mathcal{KL}(q(\bz;\phi) \| p(\bz;\eta)).
\eean
Since $\mathcal{KL}(q(\bz;\phi) \| p(\bz;\eta))$ is always non-negative, $\cL^{(2)}(\theta, \phi) \leq \cL(\theta, \phi)$ and it means that the two-step learning method with VAE \cite{dai2018diagnosing, NIPS2017_7a98af17, engel2018latent} minimizes the lower bound of \eqref{eq:stage1}.

\subsection{Step 2}

\cite{An2023DistributionalLO} showed that $\mathcal{KL}(q(\bz;\phi) \| p(\bz;\eta))$ can be minimized with the following conditions on $q(\bz;\phi)$ (note that the parameter $\phi$ is fixed during step 2):
\ben
    \item $\bz$ comprises $d$ continuous random variables,
    \item and $q(\bz; \phi)$ is defined over $\bz \in \mbR^d$ with $q(\bz; \phi) > 0$ for all $\bz \in \mbR^d$.
\een
Since our posterior distribution $q(\bz|\bx;\phi)$ is assumed to be a multivariate Gaussian distribution, it can be easily shown that the above two conditions are satisfied. 

Therefore, we can minimize $\mathcal{KL}(q(\bz;\phi) \| p(\bz;\eta))$ during the training process of step 2. Training details and relevant theorems can be found in \cite{An2023DistributionalLO}. It's also possible to model the prior distribution in step 2 using methodologies such as GMM (Gaussian Mixture Model) \cite{ijcai2017p273, Sbakan2020OnTE} or KDE (Kernel Density Estimation) to approximate the distribution of the aggregated posterior.

\subsection{Incorporating Causal Structure Information}

While not explored in this paper, assuming that the causal structure information among the observed variables is given, the need for challenging conditional independence assumptions that are hard to satisfy in high-dimensional settings can be alleviated. Recently, many studies have emerged that use continuous optimization to tackle the NP-hard problem of DAG learning \cite{zheng2018dags, yu2019dag, charpentier2022differentiable, ng2022masked, lachapelle2019gradient} and to use gradient-based optimizations such as deep learning methods. Utilizing these methods, it is possible to find a DAG (causal structure) for a given dataset.

Let $Pa(\bx_j)$ represent the parent variables of $\bx_j$ for $j=1,\cdots,p$. Additionally, let $G$ denote the graph representing the causal structure among the observed variables, and $\bx$ is the Bayesian network with respect to $G$. Then, $p(\bx)$ can be expressed as a product of individual density functions, conditional on their parent variables:
\bea \label{eq:bayesian}
p(\bx) = \prod_{j=1}^p p(\bx_j | Pa(\bx_j)).
\eea

Based on \eqref{eq:bayesian}, the lower bound on the log-likelihood of the single observation $\bx$ is written as:
\bean
\log p(\bx) = && \sum_{j=1}^p \log p(\bx_j | Pa(\bx_j)) \\
\geq && \sum_{j=1}^p \mbE_{q(\bz|\bx)} [\log p(\bx_j | \bz, Pa(\bx_j))] \\
&-& \mathcal{KL}\Big(q(\bz|\bx) \| p(\bz|Pa(\bx_j))\Big),
\eean
(for notational simplicity, we drop notations for parameters).

In the above derivation, we assumed that the posterior distribution depends on all the observed variables, but it is also possible to define the posterior distribution for each $\bx_j$ separately using only the parent variables, as follows: $q(\bz|Pa(\bx_j))$ for $j=1,\cdots,p$. Hence, when incorporating causal structure information, it's possible to define the reconstruction loss for each observed variable $\bx_j$ individually (one-dimensional distributional learning), even without relying on the conditional independence assumption.

\section{Experiments}
\label{sec:4}

\subsection{Toy Example: MNIST Dataset}
\label{sec:mnist}

We use the MNIST dataset \cite{lecun2010mnist} to illustrate the impact of the entropy regularization term on synthetic data generation performance. To examine the effect of the entropy regularization term on the latent space, we consider a 2-dimensional latent space for ease of visualization. 
The values are scaled within the range of 0 to 1, and we binarize them using a threshold of 0.5. The FID (Fréchet inception distance) score \cite{Heusel2017GANsTB} in Table \ref{tab:mnist} is computed using the MNIST test dataset and 10,000 synthetic images.

\begin{table}[ht]
\caption{MNIST dataset. `without' denotes the model is trained without the entropy regularization term, and `with' denotes the model is trained with the entropy regularization term. `avg. $\sigma^2$' is the averaged 2-dimensional posterior variance with the test dataset.}
  \centering
  \begin{tabular}{lrrrrrrrrrrrrrrrr}
    \toprule
     & without & with \\
    \midrule
avg. $\sigma^2$ & $(0.0003, 0.0005)$ & $(1.784, 1.940)$ \\
FID & 6.809 & 6.042 \\
    \bottomrule
  \end{tabular}
\label{tab:mnist}
\end{table}

\begin{figure*}[t]
    \centering
    \subfigure[Without the entropy regularization term.]{
    \includegraphics[width=0.23\linewidth]{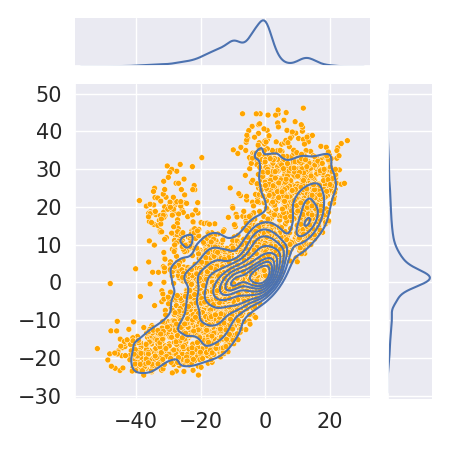}
    }
    \subfigure[With the entropy regularization term.]{
    \includegraphics[width=0.23\linewidth]{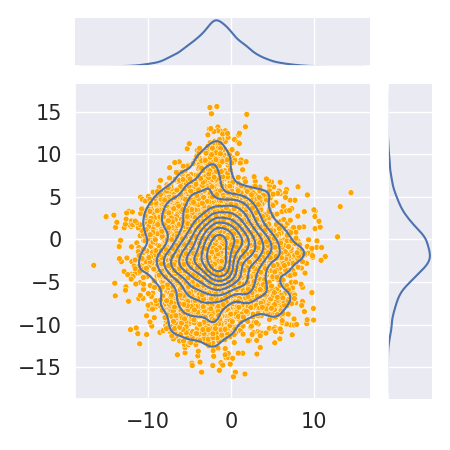}
    }
    \subfigure[Without Cramer-Wold distance regularization.]{
    \includegraphics[width=0.23\linewidth]{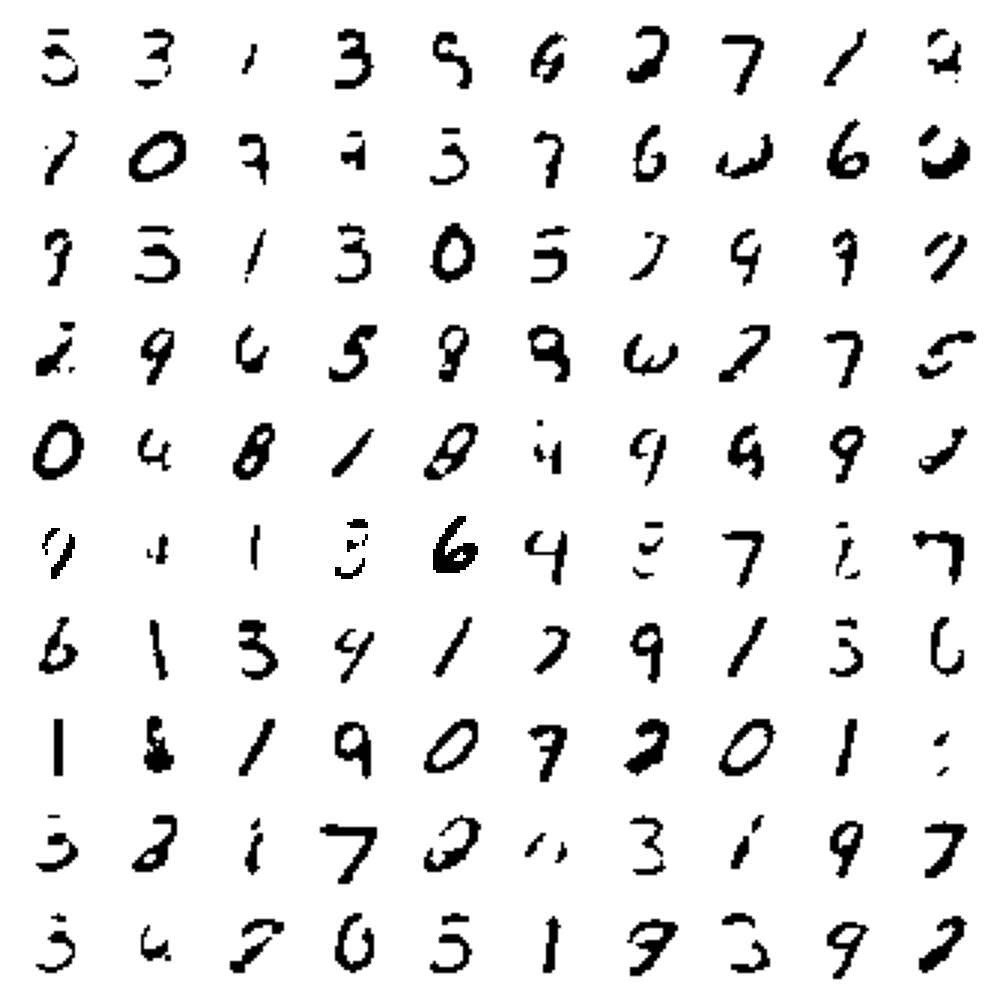}
    }
    \subfigure[With Cramer-Wold distance regularization.]{
    \includegraphics[width=0.23 \linewidth]{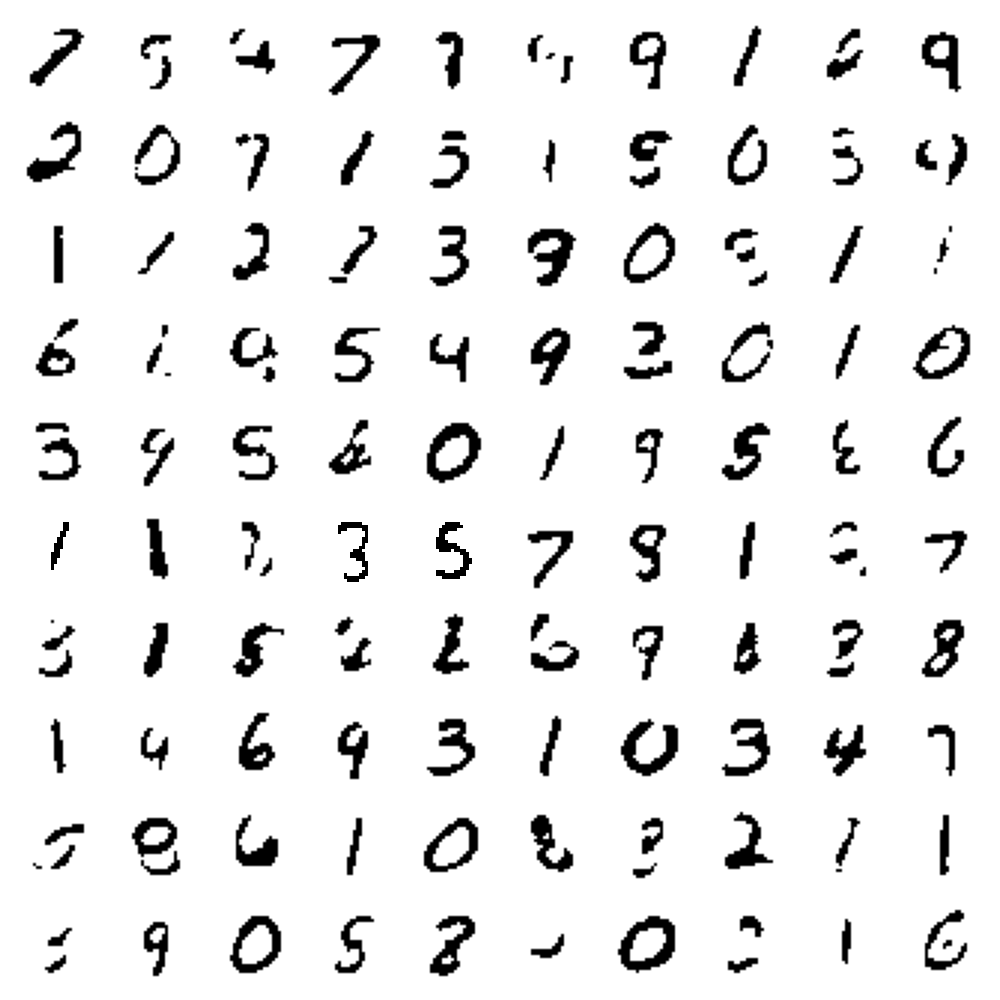}
    }
    \caption{(a)-(b) The scatter plot of sampled latent variables given the test dataset with 2-dimensional latent space (the fitted latent space).
    (c)-(d) The plot of generated samples.
    }
    \label{fig:mnist}
\end{figure*}

When examining the average 2-dimensional posterior variances in Table \ref{tab:mnist}, it becomes evident that, during the step 1 training process without the entropy regularization term, the variance of each posterior distribution is significantly smaller compared to when the entropy regularization term is present. This implies that, in the absence of the entropy regularization term, the density of each posterior distribution is concentrated in specific regions of the latent space. This, in turn, leads to increased complexity in the aggregated posterior. Also, the range of values the latent variables can take is expanded without the entropy regularization term. It can be observed by comparing the visualizations of the aggregated posterior for each test dataset with and without the entropy regularization term in Figure \ref{fig:mnist}-(a) and Figure \ref{fig:mnist}-(b).

Furthermore, when it comes to the quality of the generated MNIST images, it is evident from the FID score in Table \ref{tab:mnist} that training with the entropy regularization term leads to an increase in image quality (see Figure \ref{fig:mnist}-(c) and Figure \ref{fig:mnist}-(d)). Therefore, the entropy regularization term not only regulates the complexity of the aggregated posterior but also contributes to improving the quality of generated images. 

In short, the target distribution, represented by the aggregated posterior, becomes overly complex without the entropy regularization term, making it challenging to align the distributional learning of the aggregated posterior and the prior distribution in step 2. Consequently, a mismatch between the sampled latent variable from the prior and the aggregated posterior occurs, resulting in the generation of lower-quality synthetic data. 

\subsection{Tabular Datasets}
\label{sec:tabular}

For all experiments, the synthetic dataset is generated to have an equal number of samples as the real training dataset. We run all experiments using Geforce RTX 3090 GPU, and our experimental codes are all available with \texttt{pytorch}. We release the code at \url{XXX}. 

\subsubsection{Overview}

\textbf{Dataset.} 
We employed three real tabular datasets, all of which are characterized by high dimensionality and multiple categorical variables: \texttt{survey}, \texttt{census}, and \texttt{income}.
\texttt{survey} dataset is obtained from the 2013 American Community Survey \footnote{\url{https://www.kaggle.com/datasets/census/2013-american-community-survey?datasetId=6&sortBy=voteCount}}, which is a survey from the US Census Bureau. We sample observations corresponding to the California region (State Code: \texttt{06 .California/CA}) since the row number of the California region is the largest.
\texttt{income} dataset (Census-Income (KDD)) \footnote{\url{http://archive.ics.uci.edu/dataset/117/census+income+kdd}} is obtained from 1994 and 1995 population surveys conducted by the U.S. Census Bureau.
\texttt{census} dataset (US Census Data (1990)) \footnote{\url{https://archive.ics.uci.edu/ml/datasets/US+Census+Data+(1990)}} consists of a one percent sample of the Public Use Microdata Samples (PUMS) person records drawn from the full 1990 census sample. Due to computational issues, we sample observations from \texttt{census} and \texttt{income} randomly.  

\begin{figure*}[t]
    \centering
    \subfigure[\texttt{survey}, without]{
    \includegraphics[width=0.3 \linewidth]{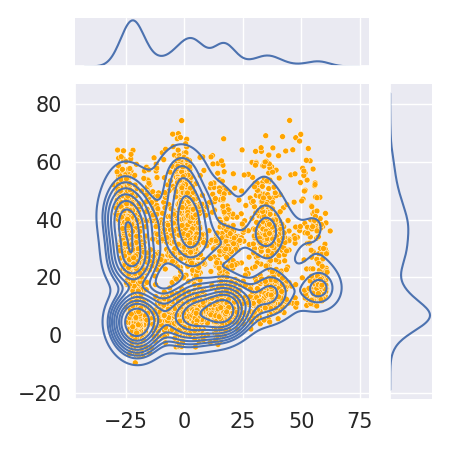}
    }
    \subfigure[\texttt{income}, without]{
    \includegraphics[width=0.3 \linewidth]{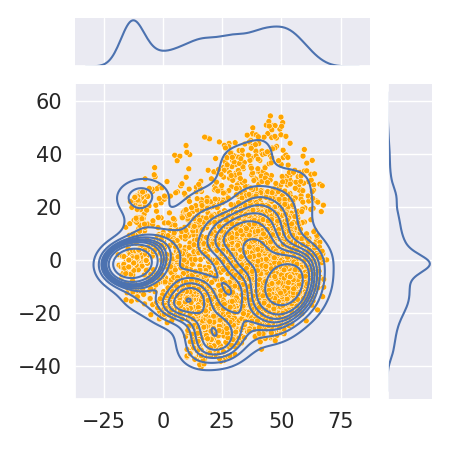}
    }
    \subfigure[\texttt{census}, without]{
    \includegraphics[width=0.3 \linewidth]{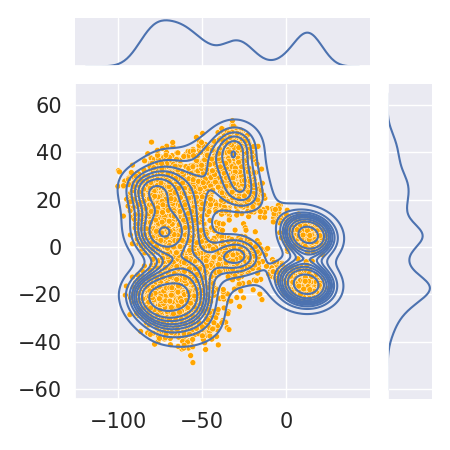}
    }
    \\
    \subfigure[\texttt{survey}, with]{
    \includegraphics[width=0.3 \linewidth]{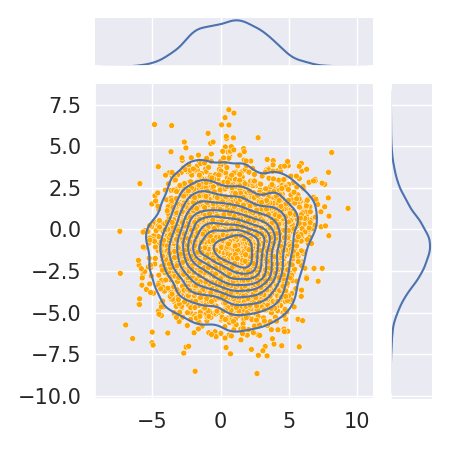}
    }
    \subfigure[\texttt{income}, with]{
    \includegraphics[width=0.3 \linewidth]{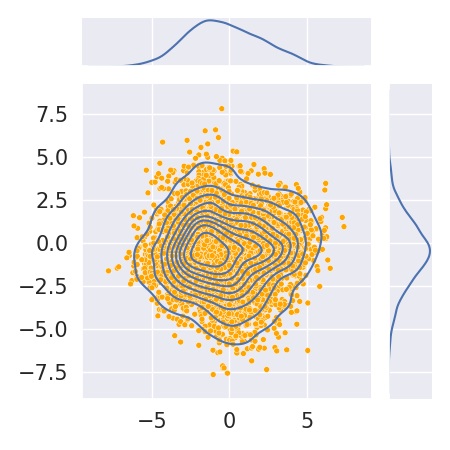}
    }
    \subfigure[\texttt{census}, with]{
    \includegraphics[width=0.3 \linewidth]{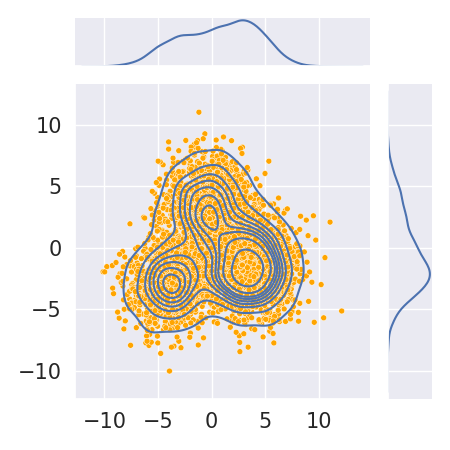}
    }
    \caption{Visualization of sampled latent variables from the aggregated posterior using dimension reduction via PCA. Top row: `without' denotes the model is trained without the entropy regularization term. Bottom row: `with' denotes the model is trained with the entropy regularization term.}
    \label{fig:tabular_latent}
\end{figure*}

\begin{table}[t]
\caption{The detailed tabular dataset descriptions. \#train and \#test denote the number of samples from the real training and test datasets, respectively. \#col indicates the number of columns, and one-hot denotes the total dimension size when each variable is transformed into one-hot vectors.}
  \centering
  \begin{tabular}{lrrrrrrrrr}
    \toprule
    dataset & \#train & \#test & \#column & one-hot \\
    \midrule
\texttt{survey} & 60,000 & 2,593 & 54 & 231 \\
\texttt{income} & 45,000 & 5,000 & 23 & 416 \\
\texttt{census} & 30,000 & 5,000 & 68 & 394 \\
    \bottomrule
  \end{tabular}
\label{tab:desc}
\end{table}

For each dataset, we tune the hyper-parameter $\beta$ and $\gamma$ differently (\texttt{survey}: $\beta = 0.05$, $\gamma = 0.5$, \texttt{income}: $\beta = 0.5$, $\gamma = 4$, \texttt{census}: $\beta = 0.1$, $\gamma = 3$). Here, $\beta$ is the weight parameter for the KL-divergence in step 2. The values range used for hyper-parameter tuning is
\bean
\lambda &\in& \{100, 500\} \\
\gamma &\in& \{0.1, 0.5, 1, 2, 3, 4, 5\} \\
\beta &\in& \{0.01, 0.05, 0.1, 0.5\}.
\eean

\textbf{Compared models.} 
We conduct a comparative analysis of our model with state-of-the-art synthesizers, which are able to generate datasets with multiple categorical variables, including both one-step and two-step learning methods. Specifically, the one-step learning methods considered are MC-Gumbel \cite{Camino2018GeneratingMS}, MC-WGAN-GP \cite{Camino2018GeneratingMS}, and WGAN-GP-A \cite{Kuo2022TheHG, Kuo2022GeneratingSC}. The two-step learning methods include medGAN \cite{Choi2017GeneratingMD}, MC-medGAN \cite{Camino2018GeneratingMS}, MC-ARAE \cite{Camino2018GeneratingMS}, corGAN \cite{Torfi2019GeneratingSH}, and DAAE \cite{Lee2020GeneratingSE}. Note that one-step learning methods involve a single training step. All the models in this comparison have a similar number of model parameters. A comprehensive comparison of the model parameters is in Table \ref{tab:num_params}.

\begin{table}[t]
\caption{The number of model parameters for each dataset.}
  \centering
  \resizebox{0.85\columnwidth}{!}{
  \begin{tabular}{lrrrrrrrrr}
    \toprule
    Model & \texttt{survey} & \texttt{income} & \texttt{census} \\
    \midrule
MC-Gumbel & 96.0K & 119.8K & 117.1K\\
MC-WGAN-GP & 96.0K & 119.8K & 117.1K\\
WGAN-GP-A & 96.0K & 119.8K & 117.1K\\
\midrule
medGAN & 96.2K & 120.1K & 117.4K\\
MC-medGAN & 96.2K & 120.1K & 117.4K\\
MC-ARAE & 92.8K & 111.4K & 109.3K\\
corGAN & 96.2K & 120.1K & 117.4K\\
DAAE & 96.1K & 120.0K & 117.3K\\
\midrule
Ours & 95.6K & 119.4K & 116.6K\\
    \bottomrule
  \end{tabular}}
\label{tab:num_params}
\end{table}

\begin{table*}[ht]
\caption{Statistical similarity results from \texttt{survey} dataset. $\uparrow$ denotes higher is better and $\downarrow$ denotes lower is better. The best value is bolded, and the second best is underlined.}
  \centering
  \resizebox{\textwidth}{!}{\begin{tabular}{lrrrrrrrrrrrrrrrr}
    \toprule
    & \multicolumn{4}{c}{marginal} & \multicolumn{4}{c}{joint} \\
    \cmidrule(lr){2-5} \cmidrule(lr){6-9}
    Model & KL $\downarrow$ & KS $\downarrow$ & Coverage $\uparrow$ & DimProb $\downarrow$ & PCD(P) $\downarrow$ & PCD(K) $\downarrow$ & log-cluster $\downarrow$ & VarPred $\downarrow$ & Rank \\
    \midrule
MC-Gumbel & $0.130_{\pm 0.120}$ & $0.106_{\pm 0.049}$ & $0.997_{\pm 0.002}$ & $1.585_{\pm 0.753}$ & $7.513_{\pm 1.397}$ & $7.360_{\pm 1.387}$ & $-2.781_{\pm 0.643}$ & $3.154_{\pm 0.082}$& 7.5\\
MC-WGAN-GP & $\underline{0.005}_{\pm 0.002}$ & $0.020_{\pm 0.003}$ & $\textbf{1.000}_{\pm 0.000}$ & $0.278_{\pm 0.051}$ & $\underline{4.585}_{\pm 0.049}$ & $\underline{4.544}_{\pm 0.044}$ & $\textbf{-5.366}_{\pm 0.357}$ & $3.054_{\pm 0.039}$& \underline{3.3}\\
WGAN-GP-A & $0.006_{\pm 0.001}$ & $\underline{0.017}_{\pm 0.003}$ & $0.875_{\pm 0.008}$ & $0.237_{\pm 0.050}$ & $nan_{\pm nan}$ & $nan_{\pm nan}$ & $\underline{-4.312}_{\pm 0.238}$ & $3.058_{\pm 0.025}$& 6.8\\
\midrule
medGAN & $0.652_{\pm 0.037}$ & $0.302_{\pm 0.013}$ & $0.999_{\pm 0.001}$ & $3.147_{\pm 0.109}$ & $7.384_{\pm 0.156}$ & $7.310_{\pm 0.159}$ & $-1.883_{\pm 0.044}$ & $1.577_{\pm 0.140}$& 8.1\\
MC-medGAN & $0.542_{\pm 0.037}$ & $0.254_{\pm 0.012}$ & $1.000_{\pm 0.000}$ & $2.631_{\pm 0.100}$ & $7.985_{\pm 0.136}$ & $7.826_{\pm 0.148}$ & $-1.842_{\pm 0.055}$ & $3.081_{\pm 0.071}$& 8.8.\\
MC-ARAE & $0.185_{\pm 0.059}$ & $0.128_{\pm 0.028}$ & $0.716_{\pm 0.157}$ & $1.769_{\pm 0.360}$ & $20.140_{\pm 0.060}$ & $20.050_{\pm 0.060}$ & $-1.878_{\pm 0.134}$ & $3.061_{\pm 0.137}$& 9.2\\
corGAN & $0.603_{\pm 0.035}$ & $0.290_{\pm 0.010}$ & $\underline{0.999}_{\pm 0.001}$ & $3.019_{\pm 0.099}$ & $7.129_{\pm 0.246}$ & $7.048_{\pm 0.243}$ & $-1.961_{\pm 0.053}$ & $\textbf{1.425}_{\pm 0.193}$& 7.2\\
DAAE & $0.258_{\pm 0.061}$ & $0.161_{\pm 0.034}$ & $0.637_{\pm 0.061}$ & $2.271_{\pm 0.362}$ & $nan_{\pm nan}$ & $nan_{\pm nan}$ & $-1.600_{\pm 0.086}$ & $3.206_{\pm 0.090}$& 10.8\\
\midrule
Ours($\lambda:0$, $\gamma:0$) & $0.128_{\pm 0.036}$ & $0.102_{\pm 0.009}$ & $0.996_{\pm 0.004}$ & $1.296_{\pm 0.148}$ & $6.742_{\pm 0.965}$ & $6.251_{\pm 0.809}$ & $-2.824_{\pm 0.172}$ & $\underline{1.516}_{\pm 0.499}$& 5.2\\
Ours($H$, $\lambda:0$, $\gamma:0$) & $0.008_{\pm 0.000}$ & $\underline{0.017}_{\pm 0.002}$ & $0.965_{\pm 0.003}$ & $0.254_{\pm 0.017}$ & $5.355_{\pm 0.260}$ & $5.316_{\pm 0.264}$ & $-2.916_{\pm 0.119}$ & $2.466_{\pm 0.749}$& 4.8\\
Ours($H$, $\lambda:100$, $\gamma:0$) & $\underline{0.005}_{\pm 0.000}$ & $\textbf{0.014}_{\pm 0.001}$ & $0.973_{\pm 0.002}$ & $\underline{0.199}_{\pm 0.013}$ & $4.629_{\pm 0.034}$ & $4.604_{\pm 0.035}$ & $-3.883_{\pm 0.176}$ & $3.022_{\pm 0.012}$& 3.6\\
Ours($H$, $\lambda:100$, $\gamma:0.5$) & $\textbf{0.004}_{\pm 0.000}$ & $\textbf{0.014}_{\pm 0.002}$ & $0.976_{\pm 0.002}$ & $\textbf{0.196}_{\pm 0.015}$ & $\textbf{4.418}_{\pm 0.301}$ & $\textbf{4.397}_{\pm 0.296}$ & $-4.037_{\pm 0.282}$ & $2.986_{\pm 0.107}$& \textbf{2.6}\\
    \bottomrule
  \end{tabular}}
\label{tab:survey}
\end{table*}

\begin{table*}[t]
\caption{Statistical similarity results from \texttt{income} dataset. $\uparrow$ denotes higher is better and $\downarrow$ denotes lower is better. The best value is bolded, and the second best is underlined.}
  \centering
  \resizebox{\textwidth}{!}{\begin{tabular}{lrrrrrrrrrrrrrrrr}
    \toprule
    & \multicolumn{4}{c}{marginal} & \multicolumn{4}{c}{joint} \\
    \cmidrule(lr){2-5} \cmidrule(lr){6-9}
    Model & KL $\downarrow$ & KS $\downarrow$ & Coverage $\uparrow$ & DimProb $\downarrow$ & PCD(P) $\downarrow$ & PCD(K) $\downarrow$ & log-cluster $\downarrow$ & VarPred $\downarrow$ & Rank \\
    \midrule
MC-Gumbel & $0.436_{\pm 0.572}$ & $0.174_{\pm 0.134}$ & $0.843_{\pm 0.213}$ & $1.465_{\pm 1.073}$ & $3.252_{\pm 1.207}$ & $3.406_{\pm 1.308}$ & $-3.069_{\pm 0.959}$ & $0.997_{\pm 0.708}$& 8.1\\
MC-WGAN-GP & $\underline{0.011}_{\pm 0.002}$ & $\underline{0.021}_{\pm 0.005}$ & $\textbf{1.000}_{\pm 0.000}$ & $\underline{0.143}_{\pm 0.040}$ & $\textbf{0.700}_{\pm 0.089}$ & $\underline{0.732}_{\pm 0.123}$ & $\underline{-5.692}_{\pm 0.340}$ & $\underline{0.129}_{\pm 0.009}$& \textbf{1.9}\\
WGAN-GP-A & $0.031_{\pm 0.003}$ & $0.027_{\pm 0.007}$ & $0.836_{\pm 0.015}$ & $0.201_{\pm 0.053}$ & $0.757_{\pm 0.227}$ & $0.749_{\pm 0.245}$ & $-5.353_{\pm 0.273}$ & $\textbf{0.127}_{\pm 0.006}$& 3.9\\
\midrule
MC-medGAN & $0.704_{\pm 0.096}$ & $0.336_{\pm 0.030}$ & $\textbf{1.000}_{\pm 0.000}$ & $2.062_{\pm 0.181}$ & $4.895_{\pm 0.391}$ & $5.043_{\pm 0.367}$ & $-2.106_{\pm 0.080}$ & $0.800_{\pm 0.119}$& 9.1\\
medGAN & $0.796_{\pm 0.173}$ & $0.351_{\pm 0.039}$ & $0.988_{\pm 0.010}$ & $2.201_{\pm 0.281}$ & $4.520_{\pm 0.457}$ & $4.657_{\pm 0.394}$ & $-2.004_{\pm 0.129}$ & $1.335_{\pm 0.348}$& 10.4\\
MC-ARAE & $0.293_{\pm 0.050}$ & $0.155_{\pm 0.016}$ & $0.412_{\pm 0.026}$ & $1.040_{\pm 0.116}$ & $nan_{\pm nan}$ & $nan_{\pm nan}$ & $-2.155_{\pm 0.081}$ & $0.839_{\pm 0.105}$& 9.2\\
corGAN & $0.639_{\pm 0.125}$ & $0.319_{\pm 0.041}$ & $0.993_{\pm 0.003}$ & $1.952_{\pm 0.192}$ & $4.247_{\pm 0.493}$ & $4.477_{\pm 0.436}$ & $-2.143_{\pm 0.083}$ & $0.966_{\pm 0.248}$& 8.5\\
DAAE & $0.351_{\pm 0.158}$ & $0.201_{\pm 0.058}$ & $0.541_{\pm 0.069}$ & $1.473_{\pm 0.409}$ & $4.741_{\pm 2.129}$ & $5.107_{\pm 2.290}$ & $-2.100_{\pm 0.332}$ & $0.762_{\pm 0.209}$& 9.5\\
\midrule
Ours($\lambda:0$, $\gamma:0$) & $0.054_{\pm 0.002}$ & $0.065_{\pm 0.003}$ & $0.881_{\pm 0.009}$ & $0.412_{\pm 0.017}$ & $2.968_{\pm 0.140}$ & $3.065_{\pm 0.173}$ & $-3.061_{\pm 0.157}$ & $0.417_{\pm 0.017}$& 6.4\\
Ours($H$, $\lambda:0$, $\gamma:0$) & $0.034_{\pm 0.005}$ & $0.046_{\pm 0.005}$ & $\underline{0.991}_{\pm 0.003}$ & $0.324_{\pm 0.025}$ & $1.012_{\pm 0.093}$ & $1.070_{\pm 0.104}$ & $-5.462_{\pm 0.349}$ & $0.195_{\pm 0.006}$& 4.5\\
Ours($H$, $\lambda:100$, $\gamma:0$) & $0.025_{\pm 0.001}$ & $0.030_{\pm 0.002}$ & $0.939_{\pm 0.008}$ & $0.225_{\pm 0.014}$ & $0.929_{\pm 0.074}$ & $0.861_{\pm 0.077}$ & $-4.674_{\pm 0.123}$ & $0.223_{\pm 0.006}$& 4.5\\
Ours($H$, $\lambda:100$, $\gamma:4$) & $\textbf{0.007}_{\pm 0.001}$ & $\textbf{0.015}_{\pm 0.003}$ & $0.942_{\pm 0.006}$ & $\textbf{0.105}_{\pm 0.019}$ & $\underline{0.730}_{\pm 0.051}$ & $\textbf{0.648}_{\pm 0.039}$ & $\textbf{-5.986}_{\pm 0.374}$ & $0.176_{\pm 0.009}$& \underline{2.0}\\
    \bottomrule
  \end{tabular}}
\label{tab:income}
\end{table*}

\begin{table*}[t]
\caption{Statistical similarity results from \texttt{census} dataset. $\uparrow$ denotes higher is better and $\downarrow$ denotes lower is better. The best value is bolded, and the second best is underlined.}
  \centering
  \resizebox{\textwidth}{!}{\begin{tabular}{lrrrrrrrrrrrrrrrr}
    \toprule
    & \multicolumn{4}{c}{marginal} & \multicolumn{4}{c}{joint} \\
    \cmidrule(lr){2-5} \cmidrule(lr){6-9}
    Model & KL $\downarrow$ & KS $\downarrow$ & Coverage $\uparrow$ & DimProb $\downarrow$ & PCD(P) $\downarrow$ & PCD(K) $\downarrow$ & log-cluster $\downarrow$ & VarPred $\downarrow$ & Rank \\
    \midrule
MC-Gumbel & $0.080_{\pm 0.032}$ & $0.079_{\pm 0.019}$ & $0.969_{\pm 0.014}$ & $1.278_{\pm 0.304}$ & $6.346_{\pm 0.940}$ & $6.291_{\pm 1.036}$ & $-3.494_{\pm 0.663}$ & $0.851_{\pm 0.138}$& 6.4\\
MC-WGAN-GP & $0.009_{\pm 0.002}$ & $0.024_{\pm 0.004}$ & $\textbf{1.000}_{\pm 0.001}$ & $0.357_{\pm 0.055}$ & $\underline{2.747}_{\pm 0.212}$ & $\textbf{2.632}_{\pm 0.190}$ & $\textbf{-5.733}_{\pm 0.495}$ & $\textbf{0.239}_{\pm 0.013}$& \textbf{2.5}\\
WGAN-GP-A & $0.016_{\pm 0.002}$ & $0.028_{\pm 0.002}$ & $0.885_{\pm 0.008}$ & $0.380_{\pm 0.042}$ & $nan_{\pm nan}$ & $nan_{\pm nan}$ & $-4.886_{\pm 0.341}$ & $\underline{0.240}_{\pm 0.014}$& 6.8\\
\midrule
MC-medGAN & $0.644_{\pm 0.044}$ & $0.302_{\pm 0.018}$ & $\textbf{1.000}_{\pm 0.000}$ & $3.249_{\pm 0.171}$ & $15.563_{\pm 0.268}$ & $16.121_{\pm 0.324}$ & $-1.678_{\pm 0.086}$ & $0.930_{\pm 0.129}$& 9.1\\
medGAN & $0.534_{\pm 0.016}$ & $0.305_{\pm 0.007}$ & $\textbf{1.000}_{\pm 0.000}$ & $3.228_{\pm 0.052}$ & $13.236_{\pm 0.599}$ & $14.128_{\pm 0.641}$ & $-1.778_{\pm 0.039}$ & $1.090_{\pm 0.085}$& 8.7\\
MC-ARAE & $0.184_{\pm 0.032}$ & $0.133_{\pm 0.018}$ & $0.505_{\pm 0.025}$ & $1.861_{\pm 0.212}$ & $nan_{\pm nan}$ & $nan_{\pm nan}$ & $-1.902_{\pm 0.180}$ & $1.230_{\pm 0.155}$& 9.6\\
corGAN & $0.471_{\pm 0.038}$ & $0.285_{\pm 0.014}$ & $\textbf{1.000}_{\pm 0.000}$ & $3.032_{\pm 0.140}$ & $12.865_{\pm 0.696}$ & $13.754_{\pm 0.681}$ & $-1.825_{\pm 0.065}$ & $1.219_{\pm 0.128}$& 8.1\\
DAAE & $0.409_{\pm 0.102}$ & $0.229_{\pm 0.042}$ & $0.541_{\pm 0.058}$ & $3.323_{\pm 0.578}$ & $nan_{\pm nan}$ & $nan_{\pm nan}$ & $-1.631_{\pm 0.068}$ & $1.983_{\pm 0.449}$& 10.9\\
\midrule
Ours($\lambda:0$, $\gamma:0$) & $0.057_{\pm 0.009}$ & $0.079_{\pm 0.008}$ & $\underline{0.998}_{\pm 0.001}$ & $1.040_{\pm 0.118}$ & $6.485_{\pm 0.633}$ & $6.776_{\pm 0.763}$ & $-3.608_{\pm 0.299}$ & $0.330_{\pm 0.029}$& 5.8\\
Ours($H$, $\lambda:0$, $\gamma:0$) & $0.010_{\pm 0.000}$ & $0.023_{\pm 0.002}$ & $0.961_{\pm 0.003}$ & $0.330_{\pm 0.024}$ & $4.831_{\pm 0.185}$ & $4.848_{\pm 0.193}$ & $-4.279_{\pm 0.383}$ & $0.487_{\pm 0.021}$& 4.6\\
Ours($H$, $\lambda:100$, $\gamma:0$) & $\underline{0.006}_{\pm 0.000}$ & $\textbf{0.020}_{\pm 0.001}$ & $0.966_{\pm 0.004}$ & $\textbf{0.269}_{\pm 0.013}$ & $3.211_{\pm 0.154}$ & $3.099_{\pm 0.144}$ & $-5.323_{\pm 0.178}$ & $0.325_{\pm 0.018}$& 3.0\\
Ours($H$, $\lambda:100$, $\gamma:3$) & $\textbf{0.005}_{\pm 0.001}$ & $\underline{0.021}_{\pm 0.002}$ & $0.953_{\pm 0.005}$ & $\underline{0.275}_{\pm 0.021}$ & $\textbf{2.744}_{\pm 0.223}$ & $\underline{2.747}_{\pm 0.240}$ & $\underline{-5.625}_{\pm 0.254}$ & $\textbf{0.239}_{\pm 0.010}$& \underline{2.6}\\
    \bottomrule
  \end{tabular}}
\label{tab:census}
\end{table*}

\begin{table*}[t]
\caption{Privacy preserving results from all tabular datasets. The lower is better. The best value is bolded, and the second best is underlined. The hyper-parameter settings for each dataset remain consistent with those used in the experiments for statistical similarity metrics.}
  \centering
  \resizebox{\textwidth}{!}{\begin{tabular}{lrrrrrrrrrrrrrrrr}
    \toprule
    & \multicolumn{3}{c}{\texttt{survey}} & \multicolumn{3}{c}{\texttt{income}} & \multicolumn{3}{c}{\texttt{census}}\\
    \cmidrule(lr){2-4} \cmidrule(lr){5-7} \cmidrule(lr){8-10}
    Model & AA(train) & AA(test) & AA(privacy) & AA(train) & AA(test) & AA(privacy) & AA(train) & AA(test) & AA(privacy)\\
    \midrule
MC-Gumbel & $0.381_{\pm 0.048}$ & $\underline{0.386}_{\pm 0.047}$ & $0.016_{\pm 0.009}$ & $0.286_{\pm 0.197}$ & $0.319_{\pm 0.154}$ & $0.043_{\pm 0.040}$ & $0.399_{\pm 0.072}$ & $0.500_{\pm 0.000}$ & $0.101_{\pm 0.072}$\\
MC-WGAN-GP & $\textbf{0.088}_{\pm 0.016}$ & $0.500_{\pm 0.000}$ & $0.588_{\pm 0.016}$ & $\textbf{0.039}_{\pm 0.010}$ & $\underline{0.221}_{\pm 0.013}$ & $0.181_{\pm 0.017}$ & $0.233_{\pm 0.005}$ & $\textbf{0.004}_{\pm 0.005}$ & $0.236_{\pm 0.004}$\\
WGAN-GP-A & $0.205_{\pm 0.017}$ & $\textbf{0.017}_{\pm 0.012}$ & $0.223_{\pm 0.014}$ & $0.106_{\pm 0.010}$ & $0.500_{\pm 0.000}$ & $0.606_{\pm 0.010}$ & $\textbf{0.054}_{\pm 0.017}$ & $\underline{0.208}_{\pm 0.008}$ & $0.154_{\pm 0.011}$\\
\midrule
medGAN & $0.416_{\pm 0.012}$ & $0.500_{\pm 0.000}$ & $0.084_{\pm 0.012}$ & $0.420_{\pm 0.048}$ & $0.467_{\pm 0.017}$ & $0.047_{\pm 0.032}$ & $0.328_{\pm 0.014}$ & $0.404_{\pm 0.008}$ & $0.076_{\pm 0.008}$\\
MC-medGAN & $0.365_{\pm 0.030}$ & $0.430_{\pm 0.020}$ & $0.065_{\pm 0.020}$ & $0.347_{\pm 0.022}$ & $0.500_{\pm 0.000}$ & $0.153_{\pm 0.022}$ & $0.354_{\pm 0.024}$ & $0.426_{\pm 0.015}$ & $0.072_{\pm 0.011}$\\
MC-ARAE & $0.489_{\pm 0.010}$ & $0.500_{\pm 0.000}$ & $\underline{0.011}_{\pm 0.010}$ & $0.415_{\pm 0.065}$ & $0.404_{\pm 0.065}$ & $\textbf{0.014}_{\pm 0.010}$ & $0.479_{\pm 0.008}$ & $0.479_{\pm 0.013}$ & $\underline{0.007}_{\pm 0.003}$\\
corGAN & $0.375_{\pm 0.054}$ & $0.444_{\pm 0.030}$ & $0.069_{\pm 0.034}$ & $0.304_{\pm 0.017}$ & $0.397_{\pm 0.012}$ & $0.093_{\pm 0.008}$ & $0.408_{\pm 0.009}$ & $0.500_{\pm 0.000}$ & $0.092_{\pm 0.009}$\\
DAAE & $0.497_{\pm 0.002}$ & $0.500_{\pm 0.000}$ & $\textbf{0.003}_{\pm 0.002}$ & $0.307_{\pm 0.079}$ & $0.333_{\pm 0.085}$ & $\underline{0.038}_{\pm 0.058}$ & $0.496_{\pm 0.003}$ & $0.491_{\pm 0.005}$ & $\textbf{0.005}_{\pm 0.005}$\\
\midrule
Ours & $\underline{0.120}_{\pm 0.003}$ & $0.500_{\pm 0.000}$ & $0.620_{\pm 0.003}$ & $\underline{0.081}_{\pm 0.020}$ & $\textbf{0.051}_{\pm 0.014}$ & $0.132_{\pm 0.013}$ & $\underline{0.063}_{\pm 0.008}$ & $0.226_{\pm 0.007}$ & $0.163_{\pm 0.006}$\\
    \bottomrule
  \end{tabular}}
\label{tab:AA}
\end{table*}

\subsubsection{Metrics 1. Statistical Similarity}

To assess the statistical similarity between the real training dataset and the synthetic data, we utilize four metrics each to measure similarity from both marginal and joint distribution perspectives. 

\textbf{Marginal.}
The marginal distributional similarity between the real training and synthetic datasets is evaluated using these four metrics: KL-divergence \cite{Goncalves2020GenerationAE, An2023DistributionalLO}, the two-sample Kolmogorov-Smirnov (KS) test \cite{Kuo2022GeneratingSC}, support coverage (category coverage) \cite{Goncalves2020GenerationAE, Kuo2022GeneratingSC}, and the MSE of dimension-wise probability \cite{Choi2017GeneratingMD, Camino2018GeneratingMS, Torfi2019GeneratingSH, cite-key}.

The \textit{KL-divergence} and \textit{KS test} are computed independently for each variable, measuring the similarity between the real training and synthetic marginal probability mass functions (PMFs). These metrics quantify the discrepancy between the two PMFs, with both being zero when the distributions are identical and larger values indicating greater dissimilarity.

The \textit{support coverage} (category coverage) metric assesses how well the synthetic data represents the support of variables in the real training data. It is computed as the mean of the ratios of the support cardinalities for all variables between the real training and synthetic datasets.
This metric is calculated as $\frac{1}{p} \sum_{j=1}^{p} T_j^* / \hat{T}_j,$
where $T_j^*$ and $\hat{T}_j$ represent the support cardinality of the $j$th variable in the real training and synthetic data, respectively. 
When the support coverage is perfect, the metric equals 1, and higher values indicate a better representation of the real data's support in the synthetic dataset.

The \textit{MSE of dimension-wise probability} measures how effectively a synthesizer has learned the distribution of the real training dataset for each dimension. It is computed as the mean squared error between the dimension-wise probability vectors for each variable in the real training and synthetic datasets.

\textbf{Joint.}
The joint distributional similarity between the real training and synthetic datasets is evaluated using these four metrics: the pairwise correlation difference using the Pearson correlation \cite{Goncalves2020GenerationAE, pmlr-v157-zhao21a, cite-key, An2023DistributionalLO}, the pairwise correlation difference using the Kendall's $\tau$ correlation \cite{Kuo2022GeneratingSC}, log-cluster \cite{Goncalves2020GenerationAE, Kuo2022GeneratingSC}, the MSE of variable-wise prediction using the multi-class classification accuracy \cite{NEURIPS2019_254ed7d2, pmlr-v157-zhao21a, Wen2021CausalTGANGT, Kamthe2021CopulaFF, Park2018DataSB, Choi2017GeneratingMD, Fang2022OvercomingCO, Camino2018GeneratingMS, Goncalves2020GenerationAE, Yang2019GroupedCG, Lee2020GeneratingSE, cite-key}.

The Pearson correlation coefficient and Kendall's $\tau$ correlation are employed to evaluate the level of correlation captured among the variables by various methods. The \textit{Pairwise Correlation Difference} (PCD) quantifies the difference in terms of the Frobenius norm between these correlation matrices calculated from the real training and synthetic datasets. A smaller PCD indicates that the synthetic data closely approximates the real data in terms of linear correlations among the variables. In essence, it assesses how effectively the method captures the linear relationships between variables present in the real dataset.

The \textit{log-cluster} metric evaluates how similar the underlying structure between the real training and synthetic datasets is, with particular attention to clustering patterns. To calculate this metric, we initially combine the real training and synthetic datasets into a unified dataset. Subsequently, we apply cluster analysis to this merged dataset using the $K$-means algorithm, using a predefined number of clusters denoted as $G$. The metric is computed as follows:
\bean
\log \left( \frac{1}{G} \sum_{i=1}^G \left( \frac{n_i^R}{n_i} - c \right)^2 \right),
\eean
where $n_i$ is the number of samples in the $i$th cluster, $n^R_i$ is the number of samples from the real dataset in the $i$th cluster, and $c = n^R / (n^R + n^S)$. $n^R$ and $n^S$ denote the number of samples from the real training and synthetic dataset. In this paper, $c$ is set to 0.5 because we have $n^R = n^S$. Large values of the log-cluster metric indicate discrepancies in cluster memberships, suggesting differences in real and synthetic data distribution. As in \cite{Goncalves2020GenerationAE}, the number of clusters is set to 20.

To assess how effectively a synthetic dataset replicates the statistical dependence structures found in the real training dataset, we utilize the \textit{MSE of variable-wise prediction} using multi-class classification accuracy. This metric is determined by evaluating the predictive performance of a trained model on both the real training and synthetic datasets. 
For each variable, we train a classifier that performs classification using all variables except the one currently under consideration (one-vs-all classifier). Due to computational issues, we use a linear logistic regression model. Subsequently, we assess the classification performance of the excluded variable on a test dataset. Finally, we calculate the MSE between the dimension-wise predictive performance (accuracy) vectors for each variable based on the classifiers trained separately on the real training and synthetic datasets. 

\subsubsection{Metrics 2. Privacy}

The privacy-preserving capacity is measured using the privacy metric proposed by \cite{Yale2020GenerationAE}. Denote $\bx_i^{(Tr)}, \bx_i^{(Te)}, \bx_i^{(S)}, i=1,\cdots,n$ as the samples from the real training, test, and synthetic datasets, respectively. And the \textit{nearest neighbor Adversarial Accuracy} (AA) between the real training and synthetic dataset is defined as:
\bean
AA_{TrS} &=& \frac{1}{2} \Bigg( \frac{1}{n} \sum_{i=1}^n \mathbbm{1}\Big(D_{TrS}(i) > D_{TrTr}(i)\Big) \\
&& + \frac{1}{n} \sum_{i=1}^n \mathbbm{1}\Big(D_{STr}(i) > D_{SS}(i)\Big) \Bigg),
\eean
where 
\bean
D_{TrS}(i) &=& \min_{j=1,\cdots,n} d(\bx_i^{(Tr)}, \bx_j^{(S)}) \\
D_{TrTr}(i) &=& \min_{j=1,\cdots,n, j \neq i} d(\bx_i^{(Tr)}, \bx_j^{(Tr)}) \\
D_{SS}(i) &=& \min_{j=1,\cdots,n, j \neq i} d(\bx_i^{(S)}, \bx_j^{(S)}),
\eean
and $\mathbbm{1}(\cdot)$ is an indicator function, $d(\cdot)$ is the Hamming distance. The AA between the real test and synthetic dataset ($AA_{TeS}$) can be defined similarly. $AA_{TrS}$ indicates the performance of an adversarial classifier responsible for distinguishing between the real training and the synthetic dataset. The ideal scenario is when $AA_{TrS}$ equals 0.5, which means that it's impossible to differentiate between the real training and synthetic datasets.

To simplify, when the generative model effectively replicates real data, the adversarial classifier finds it challenging to tell generated data apart from real data. This results in both the training and test adversarial accuracy (referred to as $AA_{TrS}$ and $AA_{TeS}$) being approximately 0.5, and the privacy loss, which is defined as $|AA_{TrS} - AA_{TeS}|$, becomes negligible. In essence, privacy is preserved.

Conversely, if the generative model performs poorly and fails to mimic real data accurately, the adversarial classifier easily distinguishes between them. Consequently, both the training and test adversarial accuracy will be high, likely exceeding 0.5, and they will have similar values. Surprisingly, even in this case, privacy loss remains low. However, the usefulness of the generated synthetic data for practical purposes may be limited.

Lastly, if the generator overfits the training data, the training adversarial accuracy will be high, indicating a good match with the training data. However, the test adversarial accuracy will hover around 0.5, indicating a poor resemblance to new, unseen data. In such a case, privacy is compromised, and the generative model struggles to generalize effectively to new data, resulting in a high privacy loss, nearing 0.5.

For ease of interpretation, Table \ref{tab:AA} reports the differences between $AA_{TrS}$, $AA_{TeS}$, and 0.5. And AA(train), AA(test), and AA(privacy) represent $|AA_{TrS}-0.5|$, $|AA_{TeS}-0.5|$, and $|AA_{TrS} - AA_{TeS}|$, respectively.

\subsubsection{Results}

\textbf{Metrics 1. Statistical Similarity.} In Tables \ref{tab:survey}, \ref{tab:income}, and \ref{tab:census}, we empirically assess how our proposed model performs in comparison to baseline models (both one-step and two-step learning methods) in terms of metrics that measure statistical similarity from both marginal and joint distribution perspectives.

In the following result tables, we abbreviate the metrics as follows: KL-divergence as KL, the two-sample Kolmogorov-Smirnov test as KS, support coverage (category coverage) as Coverage, MSE of dimension-wise probability as DimProb, pairwise correlation difference using Pearson correlation as PCD(P), pairwise correlation difference using Kendall's $\tau$ correlation as PCD(K), log-cluster as log-cluster, and MSE of variable-wise prediction using multi-class classification accuracy as VarPred. Lastly, the Rank column represents the ranking of each model based on each metric, with the average rank computed for each model across all metrics.

Firstly, to assess the impact of regularization terms, we performed an ablation study by training the model with and without the entropy regularization term, Cramer-Wold distance, and classification loss regularization. In Tables \ref{tab:survey}, \ref{tab:income}, and \ref{tab:census}, $H$ indicates the inclusion of the entropy regularization term, and $\lambda$ and $\gamma$ represent the weight parameters for the Cramer-Wold distance and the classification loss regularization used in \eqref{eq:final_obj}. In other words, positive values of these weight parameters mean the incorporation of the respective regularization during training. Note that the training process of step 1 is equivalent to training a vanilla AutoEncoder without the entropy regularization term.

Tables \ref{tab:survey}, \ref{tab:income}, and \ref{tab:census} consistently demonstrate that as we add the entropy regularization term, the Cramer-Wold distance, and the classification loss regularization, the average rank progressively increases. This suggests that the entropy regularization term and the introduced regularization techniques in this paper are effective for both marginal and joint distributional learning, enhancing synthetic data generation performance. 

In particular, similar to the experimental results on the MNIST dataset in Section \ref{sec:mnist}, it can be observed that using the entropy regularization term, which was not considered in many existing two-step learning methods, helps prevent the step 2 target distribution, the aggregated posterior, from becoming overly complex. This, in turn, contributes to improving synthetic data generation performance. 

The visualization results of sampled latent variables from the aggregated posterior for each dataset can be observed in Figure \ref{fig:tabular_latent}. As shown in Figure \ref{fig:tabular_latent}, when the model is trained in step 1 without the entropy regularization term, we observe that the aggregated posterior exhibits a high level of complexity. Furthermore, in the absence of the entropy regularization term, the range of values that the latent variables can take is significantly expanded. In the visualization process, a dimension reduction technique PCA (Principal Component Analysis) is utilized to preserve the global structure of the sampled latent variables.

Table \ref{tab:survey} demonstrates that our proposed model attains the highest average rank when evaluated on \texttt{survey} dataset. This indicates that our model generates synthetic data with the most effective performance. Furthermore, in comparison to the top-performing alternative model, MC-WGAN-GP, our model outperforms it by achieving the top scores on 5 out of 8 metrics, consistently showcasing superior performance (MC-WGAP-GP achieves the top scores on 2 out of 8 metrics).

When analyzing Table \ref{tab:income}, we observe that our proposed model, while having a slightly lower average rank compared to the top-performing alternative model, MC-WGAN-GP, excels when examining the metrics where the highest scores are achieved. MC-WGAN-GP outperforms in only 2 out of 8 metrics, whereas our proposed model secures the highest scores in 5 out of 8 metrics. This consistent performance across a range of metrics underscores the strong competitiveness of our approach to generating synthetic data.

In the case of \texttt{census} dataset (Table \ref{tab:census}), our proposed model exhibits a very slight decrease in average rank compared to the top-performing alternative model, MC-WGAN-GP. Additionally, it achieves the highest metric score in 3 out of 8 metrics, which is slightly less than MC-WGAN-GP's 4 out of 8. However, our model secures either 1st or 2nd place in all metrics except the Coverage metric. Notably, in the marginal statistical similarity metrics, our model outperforms MC-WGAN-GP significantly. Thus, while our model's rank is lower in the Coverage metric, resulting in a lower average rank, it demonstrates competitive or superior performance in the other seven metrics.

\textbf{Metrics 2. Privacy.} Comparing AA(train) in Table \ref{tab:AA} with MC-WGAN-GP, the top-performing model among the statistical similarity metrics, our proposed model consistently shows competitive results, being the second-best performer in \texttt{survey} and \texttt{income} datasets. Notably, it outperforms MC-WGAN-GP by a significant margin in \texttt{census} dataset. This indicates that our model effectively replicates real data, making the generated data practically useful and challenging for the adversarial classifier to distinguish from real training data. In terms of AA(test) in Table \ref{tab:AA}, which represents the ability to reproduce results for unseen data, our model exhibits competitive performance across all datasets. Moreover, except for \texttt{survey} dataset, our model shows minimal differences between AA(train) and AA(test) compared to other models, indicating that it doesn't overfit the real training dataset.

Regarding privacy-preserving performance as shown by AA(privacy) in Table \ref{tab:AA}, our proposed model, when compared to MC-WGAN-GP, demonstrates lower values in \texttt{income} and \texttt{census} datasets, and competitive values in \texttt{survey} dataset. Hence, it can be concluded that our proposed model carries a lower risk of privacy leakage.

\section{Conclusion and Limitations}
\label{sec:5}

This paper introduces a novel two-step learning framework designed to estimate both the marginal and joint distributions effectively. We achieve this by incorporating two regularization techniques: the Cramer-Wold distance and the classification loss. Moreover, we establish a theoretical distinction between our proposed approach and existing two-step learning methods. Specifically, we provide an accurate mathematical derivation of the objective function in step 1, introducing the previously overlooked entropy regularization term (the second term in \eqref{eq:stage1} on the right-hand side). We demonstrate that this term plays a significant role in enhancing the synthetic data generation performance of our model, distinguishing it from conventional two-step learning.

For all datasets, the parameter $\lambda$ consistently demonstrates good performance without the need for tuning, with a fixed value of 100. However, the weight parameter for the classification loss regularization and $\beta$, the weight parameter for KL-divergence in step 2, require dataset-specific tuning. Additionally, while our approach exhibits strong performance in most metrics measuring statistical similarity with the marginal distribution, it falls short in terms of support coverage compared to other models. Addressing this limitation will be a part of our future work.

On the other hand, the introduction of Electronic Health Records (EHR) has led to the generation of vast amounts of data, and numerous studies have focused on the generation of synthetic EHRs to facilitate data-driven research while addressing privacy concerns and the risk of re-identification \cite{Choi2017GeneratingMD, Goncalves2020GenerationAE, Park2021SynthesizingIC, Kuo2022GeneratingSC, Yang2019GroupedCG, Lu2022MultiLabelCT, cite-key, Kuo2022TheHG}. Since many EHR datasets consist of discrete or categorical variables, we anticipate that our proposed methodology can be readily applied to synthetic data generation for EHR data.

In this regard, our paper primarily focuses on improving synthetic data generation performance. And, due to the trade-off between the quality of synthetic data and privacy preservation, our approach provides a lower level of privacy protection than other models. Hence, our future work will involve refining our proposed method to enhance both synthetic data generation for EHR data and the model's privacy-preserving performance concurrently.



\appendix
\section{Appendix}

\subsection{Mathematical Derivations}
\label{app:twostep}

\subsubsection{The forward KL-divergence}
\bean
&& \mathcal{KL}(p(\bx) \| \hat{p}(\bx)) \\
&=& \mathcal{KL}\left( p(\bx) \| \int p(\bz) p(\bx|\bz) d\bz \right) \\
&=& \int p(\bx) \log p(\bx) d\bx - \mbE_{p(\bx)} \left[ \log \int p(\bz) p(\bx|\bz) d\bz \right] \\
&=& -H (p(\bx)) - \mbE_{p(\bx)} \left[ \log \int p(\bz) p(\bx|\bz) \frac{q(\bz|\bx)}
{q(\bz|\bx)} d\bz \right] \\
&\leq& -H (p(\bx)) - \mbE_{p(\bx)} \left[ \int q(\bz|\bx) \log p(\bx|\bz) \frac{p(\bz)}{q(\bz|\bx)} d\bz \right] \\
&=& \int p(\bx) q(\bz|\bx) \log p(\bx) d\bx d\bz \\
&& - \int p(\bx) q(\bz|\bx) \log \frac{p(\bx|\bz) p(\bz)}{q(\bz|\bx)} d\bz \\
&=& \int p(\bx) q(\bz|\bx) \log \frac{p(\bx) q(\bz|\bx)}{p(\bx|\bz) p(\bz)} \cdot \frac{q(\bz)}{q(\bz)} d\bz \\
&=& \int p(\bx) q(\bz|\bx) \log \frac{p(\bx) q(\bz|\bx)}{p(\bx|\bz) q(\bz)} d\bz \\
&& + \int p(\bx) q(\bz|\bx) \log \frac{q(\bz)}{p(\bz)} d\bz \\
&=& \underbrace{\mathcal{KL}\Big(p(\bx) q(\bz|\bx) \| p(\bx|\bz) q(\bz)\Big)}_{(i)} + \underbrace{\mathcal{KL}\Big(q(\bz) \| p(\bz)\Big)}_{(ii)}
\eean
where $H(\cdot)$ is the entropy function (for notational simplicity, we drop notations for parameters).

\subsubsection{The upper bound of the entropy regularization term}
\bean
&& \mbE_{p(\bx)} [\mathcal{KL}(q(\bz|\bx;\phi) \| q(\bz;\phi))] \\
&=& \mbE_{p(\bx) q(\bz|\bx;\phi)} 
[\log q(\bz|\bx;\phi)] - \mbE_{p(\bx) q(\bz|\bx;\phi)} [\log q(\bz;\phi)],
\eean
and
\bean
&& \mbE_{p(\bx) q(\bz|\bx;\phi)} [\log q(\bz;\phi)] \\
&=& \int q(\bz;\phi) \log q(\bz;\phi) d\bz \\
&=& \int q(\bz;\phi) \log \left( \int p(\bx) q(\bz|\bx;\phi) d\bx \right) d\bz \\
&\geq& \iint p(\bx) q(\bz;\phi) \log q(\bz|\bx;\phi) d\bx d\bz.
\eean
Therefore, 
\bean
&& \mbE_{p(\bx)} [\mathcal{KL}(q(\bz|\bx;\phi) \| q(\bz;\phi))] \\
&\leq& \mbE_{p(\bx) q(\bz|\bx;\phi)} 
[\log q(\bz|\bx;\phi)] - \mbE_{p(\bx) q(\bz;\phi)} 
[\log q(\bz|\bx;\phi)].
\eean

\bibliographystyle{plain}
\bibliography{ref}

\begin{thebibliography}{10}

\bibitem{pmlr-v80-achlioptas18a}
Panos Achlioptas, Olga Diamanti, Ioannis Mitliagkas, and Leonidas Guibas.
\newblock Learning representations and generative models for 3{D} point clouds.
\newblock In Jennifer Dy and Andreas Krause, editors, {\em Proceedings of the
  35th International Conference on Machine Learning}, volume~80 of {\em
  Proceedings of Machine Learning Research}, pages 40--49. PMLR, 10--15 Jul
  2018.

\bibitem{An2023DistributionalLO}
Seunghwan An and Jong-June Jeon.
\newblock Distributional learning of variational autoencoder: Application to
  synthetic data generation.
\newblock 2023.

\bibitem{pmlr-v70-arjovsky17a}
Martin Arjovsky, Soumith Chintala, and L{\'e}on Bottou.
\newblock {W}asserstein generative adversarial networks.
\newblock In Doina Precup and Yee~Whye Teh, editors, {\em Proceedings of the
  34th International Conference on Machine Learning}, volume~70 of {\em
  Proceedings of Machine Learning Research}, pages 214--223. PMLR, 06--11 Aug
  2017.

\bibitem{Blei2016VariationalIA}
David~M. Blei, Alp Kucukelbir, and Jon~D. McAuliffe.
\newblock Variational inference: A review for statisticians.
\newblock {\em Journal of the American Statistical Association}, 112:859 --
  877, 2016.

\bibitem{Bousquet2017FromOT}
Olivier Bousquet, Sylvain Gelly, Ilya~O. Tolstikhin, Carl-Johann Simon-Gabriel,
  and Bernhard Schoelkopf.
\newblock From optimal transport to generative modeling: the vegan cookbook.
\newblock {\em arXiv: Machine Learning}, 2017.

\bibitem{Camino2018GeneratingMS}
Ramiro~Daniel Camino, Christian~A. Hammerschmidt, and Radu State.
\newblock Generating multi-categorical samples with generative adversarial
  networks.
\newblock {\em ArXiv}, abs/1807.01202, 2018.

\bibitem{charpentier2022differentiable}
Bertrand Charpentier, Simon Kibler, and Stephan G{\"u}nnemann.
\newblock Differentiable {DAG} sampling.
\newblock In {\em International Conference on Learning Representations}, 2022.

\bibitem{Choi2017GeneratingMD}
E.~Choi, Siddharth Biswal, Bradley~A. Malin, Jon~D. Duke, Walter~F. Stewart,
  and Jimeng Sun.
\newblock Generating multi-label discrete patient records using generative
  adversarial networks.
\newblock In {\em Machine Learning in Health Care}, 2017.

\bibitem{dai2018diagnosing}
Bin Dai and David Wipf.
\newblock Diagnosing and enhancing {VAE} models.
\newblock In {\em International Conference on Learning Representations}, 2019.

\bibitem{Lu2022MultiLabelCT}
Chang de~Lu, Chandan~K. Reddy, Ping Wang, Dong Nie, and Yue Ning.
\newblock Multi-label clinical time-series generation via conditional gan.
\newblock {\em ArXiv}, abs/2204.04797, 2022.

\bibitem{Deshpande2018GenerativeMU}
Ishani Deshpande, Ziyu Zhang, and Alexander~G. Schwing.
\newblock Generative modeling using the sliced wasserstein distance.
\newblock {\em 2018 IEEE/CVF Conference on Computer Vision and Pattern
  Recognition}, pages 3483--3491, 2018.

\bibitem{10.5555/3020847.3020875}
Gintare~Karolina Dziugaite, Daniel~M. Roy, and Zoubin Ghahramani.
\newblock Training generative neural networks via maximum mean discrepancy
  optimization.
\newblock In {\em Proceedings of the Thirty-First Conference on Uncertainty in
  Artificial Intelligence}, UAI'15, page 258–267, Arlington, Virginia, USA,
  2015. AUAI Press.

\bibitem{engel2018latent}
Jesse Engel, Matthew Hoffman, and Adam Roberts.
\newblock Latent constraints: Learning to generate conditionally from
  unconditional generative models.
\newblock In {\em International Conference on Learning Representations}, 2018.

\bibitem{Fang2022OvercomingCO}
Kevin Fang, Vaikkunth Mugunthan, Vayd Ramkumar, and Lalana Kagal.
\newblock Overcoming challenges of synthetic data generation.
\newblock {\em 2022 IEEE International Conference on Big Data (Big Data)},
  pages 262--270, 2022.

\bibitem{Goncalves2020GenerationAE}
Andre Goncalves, Priyadip Ray, Braden~C. Soper, Jennifer~L. Stevens, Linda
  Coyle, and Ana~Paula Sales.
\newblock Generation and evaluation of synthetic patient data.
\newblock {\em BMC Medical Research Methodology}, 20, 2020.

\bibitem{guo2020variational}
Chunsheng Guo, Jialuo Zhou, Huahua Chen, Na~Ying, Jianwu Zhang, and Di~Zhou.
\newblock Variational autoencoder with optimizing gaussian mixture model
  priors.
\newblock {\em IEEE Access}, 8:43992--44005, 2020.

\bibitem{hajimiri2021semi}
Sina Hajimiri, Aryo Lotfi, and Mahdieh~Soleymani Baghshah.
\newblock Semi-supervised disentanglement of class-related and
  class-independent factors in vae.
\newblock {\em arXiv preprint arXiv:2102.00892}, 2021.

\bibitem{Hernandez2022SyntheticDG}
Mikel Hernandez, Gorka Epelde, Ane Alberdi, Rodrigo Cilla, and Debbie Rankin.
\newblock Synthetic data generation for tabular health records: A systematic
  review.
\newblock {\em Neurocomputing}, 493:28--45, 2022.

\bibitem{Heusel2017GANsTB}
Martin Heusel, Hubert Ramsauer, Thomas Unterthiner, Bernhard Nessler, and Sepp
  Hochreiter.
\newblock Gans trained by a two time-scale update rule converge to a local nash
  equilibrium.
\newblock In {\em Neural Information Processing Systems}, 2017.

\bibitem{Hofert2021RafterNetPP}
Marius Hofert, Avinash Prasad, and Mu~Zhu.
\newblock Rafternet: Probabilistic predictions in multi-response regression.
\newblock {\em ArXiv}, abs/2112.03377, 2021.

\bibitem{hoffman2016elbo}
Matthew~D Hoffman and Matthew~J Johnson.
\newblock Elbo surgery: yet another way to carve up the variational evidence
  lower bound.
\newblock In {\em Workshop in Advances in Approximate Bayesian Inference,
  NIPS}, volume~1, 2016.

\bibitem{jang2017categorical}
Eric Jang, Shixiang Gu, and Ben Poole.
\newblock Categorical reparameterization with gumbel-softmax.
\newblock In {\em International Conference on Learning Representations}, 2017.

\bibitem{ijcai2017p273}
Zhuxi Jiang, Yin Zheng, Huachun Tan, Bangsheng Tang, and Hanning Zhou.
\newblock Variational deep embedding: An unsupervised and generative approach
  to clustering.
\newblock In {\em Proceedings of the Twenty-Sixth International Joint
  Conference on Artificial Intelligence, {IJCAI-17}}, pages 1965--1972, 2017.

\bibitem{Kamthe2021CopulaFF}
Sanket Kamthe, Samuel~A. Assefa, and Marc~Peter Deisenroth.
\newblock Copula flows for synthetic data generation.
\newblock {\em ArXiv}, abs/2101.00598, 2021.

\bibitem{Kingma2014}
Diederik~P. Kingma and Max Welling.
\newblock {Auto-Encoding Variational Bayes}.
\newblock In {\em 2nd International Conference on Learning Representations,
  {ICLR} 2014, Banff, AB, Canada, April 14-16, 2014, Conference Track
  Proceedings}, 2014.

\bibitem{KNOP2022119}
Szymon Knop, Marcin Mazur, Przemys{\l}aw Spurek, Jacek Tabor, and Igor Podolak.
\newblock Generative models with kernel distance in data space.
\newblock {\em Neurocomputing}, 487:119--129, 2022.

\bibitem{Kuo2022GeneratingSC}
Nicholas I-Hsien Kuo, Louisa~R Jorm, and Sebastiano Barbieri.
\newblock Generating synthetic clinical data that capture class imbalanced
  distributions with generative adversarial networks: Example using
  antiretroviral therapy for hiv.
\newblock {\em ArXiv}, abs/2208.08655, 2022.

\bibitem{Kuo2022TheHG}
Nicholas I-Hsien Kuo, Mark~N. Polizzotto, Simon Finfer, Federico Garcia, Anders
  S{\"o}nnerborg, Maurizio Zazzi, Michael B{\"o}hm, Louisa~R Jorm, and
  Sebastiano Barbieri.
\newblock The health gym: synthetic health-related datasets for the development
  of reinforcement learning algorithms.
\newblock {\em Scientific Data}, 9, 2022.

\bibitem{lachapelle2019gradient}
S{\'e}bastien Lachapelle, Philippe Brouillard, Tristan Deleu, and Simon
  Lacoste-Julien.
\newblock Gradient-based neural dag learning.
\newblock {\em arXiv preprint arXiv:1906.02226}, 2019.

\bibitem{lecun2010mnist}
Yann LeCun, Corinna Cortes, and CJ~Burges.
\newblock Mnist handwritten digit database.
\newblock {\em ATT Labs [Online]. Available: http://yann.lecun.com/exdb/mnist},
  2, 2010.

\bibitem{Lee2020GeneratingSE}
Dongha Lee, Hwanjo Yu, Xiaoqian Jiang, Deevakar Rogith, Meghana Gudala, Mubeen
  Tejani, Qiuchen Zhang, and Li~Xiong.
\newblock Generating sequential electronic health records using dual
  adversarial autoencoder.
\newblock {\em Journal of the American Medical Informatics Association :
  JAMIA}, 27 9:1411--1419, 2020.

\bibitem{cite-key}
Jin Li, Benjamin~J. Cairns, Jingsong Li, and Tingting Zhu.
\newblock Generating synthetic mixed-type longitudinal electronic health
  records for artificial intelligent applications.
\newblock {\em npj Digital Medicine}, 6(1):98, 2023.

\bibitem{Li2015GenerativeMM}
Yujia Li, Kevin Swersky, and Richard~S. Zemel.
\newblock Generative moment matching networks.
\newblock In {\em International Conference on Machine Learning}, 2015.

\bibitem{Makhzani2015AdversarialA}
Alireza Makhzani, Jonathon Shlens, Navdeep Jaitly, and Ian~J. Goodfellow.
\newblock Adversarial autoencoders.
\newblock {\em ArXiv}, abs/1511.05644, 2015.

\bibitem{ng2022masked}
Ignavier Ng, Shengyu Zhu, Zhuangyan Fang, Haoyang Li, Zhitang Chen, and Jun
  Wang.
\newblock Masked gradient-based causal structure learning.
\newblock In {\em Proceedings of the 2022 SIAM International Conference on Data
  Mining (SDM)}, pages 424--432. SIAM, 2022.

\bibitem{Papamakarios2019NormalizingFF}
George Papamakarios, Eric~T. Nalisnick, Danilo~Jimenez Rezende, Shakir Mohamed,
  and Balaji Lakshminarayanan.
\newblock Normalizing flows for probabilistic modeling and inference.
\newblock {\em J. Mach. Learn. Res.}, 22:57:1--57:64, 2019.

\bibitem{Park2021SynthesizingIC}
Nari Park, Yeong~Hyeon Gu, and Seong~Joon Yoo.
\newblock Synthesizing individual consumers' credit historical data using
  generative adversarial networks.
\newblock {\em Applied Sciences}, 2021.

\bibitem{Park2018DataSB}
Noseong Park, Mahmoud Mohammadi, Kshitij Gorde, Sushil Jajodia, Hongkyu Park,
  and Youngmin Kim.
\newblock Data synthesis based on generative adversarial networks.
\newblock {\em Proc. VLDB Endow.}, 11:1071--1083, 2018.

\bibitem{JimenezRezende2014StochasticBA}
Danilo~Jimenez Rezende, Shakir Mohamed, and Daan Wierstra.
\newblock Stochastic backpropagation and approximate inference in deep
  generative models.
\newblock In {\em International Conference on Machine Learning}, 2014.

\bibitem{Sbakan2020OnTE}
Yusuf~Cem S{\"u}bakan, Maxime Gasse, and Laurent Charlin.
\newblock On the effectiveness of two-step learning for latent-variable models.
\newblock {\em 2020 IEEE 30th International Workshop on Machine Learning for
  Signal Processing (MLSP)}, pages 1--6, 2020.

\bibitem{Tabor2018CramerWoldA}
Jacek Tabor, Szymon Knop, Przemyslaw Spurek, Igor~T. Podolak, Marcin Mazur, and
  Stanislaw Jastrzebski.
\newblock Cramer-wold autoencoder.
\newblock {\em J. Mach. Learn. Res.}, 21:164:1--164:28, 2018.

\bibitem{takahashi2019variational}
Hiroshi Takahashi, Tomoharu Iwata, Yuki Yamanaka, Masanori Yamada, and Satoshi
  Yagi.
\newblock Variational autoencoder with implicit optimal priors.
\newblock In {\em Proceedings of the AAAI Conference on Artificial
  Intelligence}, volume~33, pages 5066--5073, 2019.

\bibitem{Tolstikhin2017WassersteinA}
Ilya~O. Tolstikhin, Olivier Bousquet, Sylvain Gelly, and Bernhard
  Sch{\"o}lkopf.
\newblock Wasserstein auto-encoders.
\newblock {\em ArXiv}, abs/1711.01558, 2017.

\bibitem{pmlr-v84-tomczak18a}
Jakub Tomczak and Max Welling.
\newblock Vae with a vampprior.
\newblock In Amos Storkey and Fernando Perez-Cruz, editors, {\em Proceedings of
  the Twenty-First International Conference on Artificial Intelligence and
  Statistics}, volume~84 of {\em Proceedings of Machine Learning Research},
  pages 1214--1223. PMLR, 09--11 Apr 2018.

\bibitem{tomczak2018vae}
Jakub Tomczak and Max Welling.
\newblock Vae with a vampprior.
\newblock In {\em International Conference on Artificial Intelligence and
  Statistics}, pages 1214--1223. PMLR, 2018.

\bibitem{Torfi2019GeneratingSH}
Amirsina Torfi and Mohammadreza Beyki.
\newblock Generating synthetic healthcare records using convolutional
  generative adversarial networks.
\newblock 2019.

\bibitem{NIPS2017_7a98af17}
Aaron van~den Oord, Oriol Vinyals, and koray kavukcuoglu.
\newblock Neural discrete representation learning.
\newblock In I.~Guyon, U.~Von Luxburg, S.~Bengio, H.~Wallach, R.~Fergus,
  S.~Vishwanathan, and R.~Garnett, editors, {\em Advances in Neural Information
  Processing Systems}, volume~30. Curran Associates, Inc., 2017.

\bibitem{Wen2021CausalTGANGT}
Bingyang Wen, Luis~Oliveros Colon, K.P. Subbalakshmi, and Ramamurti
  Chandramouli.
\newblock Causal-tgan: Generating tabular data using causal generative
  adversarial networks.
\newblock {\em ArXiv}, abs/2104.10680, 2021.

\bibitem{NEURIPS2019_254ed7d2}
Lei Xu, Maria Skoularidou, Alfredo Cuesta-Infante, and Kalyan Veeramachaneni.
\newblock Modeling tabular data using conditional gan.
\newblock In H.~Wallach, H.~Larochelle, A.~Beygelzimer, F.~d\textquotesingle
  Alch\'{e}-Buc, E.~Fox, and R.~Garnett, editors, {\em Advances in Neural
  Information Processing Systems}, volume~32. Curran Associates, Inc., 2019.

\bibitem{Yale2020GenerationAE}
Andrew Yale, Saloni Dash, Ritik Dutta, Isabelle~M Guyon, Adrien Pavao, and
  Kristin~P. Bennett.
\newblock Generation and evaluation of privacy preserving synthetic health
  data.
\newblock {\em Neurocomputing}, 416:244--255, 2020.

\bibitem{Yang2019GroupedCG}
Fan Yang, Zhongping Yu, Yunfan Liang, Xiaolu Gan, Kaibiao Lin, Quan Zou, and
  Yifeng Zeng.
\newblock Grouped correlational generative adversarial networks for discrete
  electronic health records.
\newblock {\em 2019 IEEE International Conference on Bioinformatics and
  Biomedicine (BIBM)}, pages 906--913, 2019.

\bibitem{yu2019dag}
Yue Yu, Jie Chen, Tian Gao, and Mo~Yu.
\newblock Dag-gnn: Dag structure learning with graph neural networks.
\newblock In {\em International Conference on Machine Learning}, pages
  7154--7163. PMLR, 2019.

\bibitem{Zhao2017AdversariallyRA}
Junbo~Jake Zhao, Yoon Kim, Kelly~W. Zhang, Alexander~M. Rush, and Yann LeCun.
\newblock Adversarially regularized autoencoders.
\newblock In {\em International Conference on Machine Learning}, 2017.

\bibitem{pmlr-v157-zhao21a}
Zilong Zhao, Aditya Kunar, Robert Birke, and Lydia~Y. Chen.
\newblock Ctab-gan: Effective table data synthesizing.
\newblock In Vineeth~N. Balasubramanian and Ivor Tsang, editors, {\em
  Proceedings of The 13th Asian Conference on Machine Learning}, volume 157 of
  {\em Proceedings of Machine Learning Research}, pages 97--112. PMLR, 17--19
  Nov 2021.

\bibitem{zheng2018dags}
Xun Zheng, Bryon Aragam, Pradeep~K Ravikumar, and Eric~P Xing.
\newblock Dags with no tears: Continuous optimization for structure learning.
\newblock {\em Advances in Neural Information Processing Systems}, 31, 2018.

\end{thebibliography}

\end{document}